\documentclass[nohyperref]{article}

\usepackage{epsfig}
\usepackage{graphicx}
\usepackage{amsmath}
\usepackage{amssymb}

\usepackage{enumerate}
\usepackage{multirow}
\usepackage{algorithm}
\usepackage{algorithmic}
\usepackage{tabularx}
\usepackage{xcolor,colortbl}
\usepackage{nicefrac}
\usepackage{booktabs}
\usepackage{makecell}

\usepackage{hyperref}
\usepackage{url}

\usepackage[accepted]{icml2022}

\newtheorem{proposition}{Proposition}[section]
\newtheorem{theorem}{Theorem}[section]

\def\R{\mathbb{R}}

\newcommand{\norm}[1]{\left\|#1\right\|}

\def\D{\mathcal{D}}

\def\minop{\mathop{\rm min}\limits}
\def\maxop{\mathop{\rm max}\limits}

\def\max{\mathop{\rm max}\nolimits}

\newcolumntype{C}[1]{>{\centering\arraybackslash}p{#1}}
\newcolumntype{L}[1]{>{\raggedright\arraybackslash}p{#1}}
\newcolumntype{R}[1]{>{\raggedleft\arraybackslash}p{#1}}

\newcommand{\MH}[1]{{\color{black} #1}}
\newcommand{\MHREV}[1]{{\color{black} #1}}

\definecolor{Gray}{gray}{0.85}
\definecolor{LightCyan}{rgb}{0.88,1,1}

\newcommand{\hlrev}[1]{\textcolor{black}{#1}}

\begin{document}

\twocolumn[
 \icmltitle{
Adversarial Robustness against Multiple and Single $l_p$-Threat Models via Quick Fine-Tuning of Robust Classifiers
}

\icmlsetsymbol{equal}{*}

\begin{icmlauthorlist}
\icmlauthor{Francesco Croce}{tue}
\icmlauthor{Matthias Hein}{tue}

\end{icmlauthorlist}

\icmlaffiliation{tue}{University of T{\"u}bingen, Germany}

\icmlcorrespondingauthor{Francesco Croce}{francesco.croce@uni-tuebingen.de}

\icmlkeywords{Machine Learning, ICML}

\vskip 0.3in
]

\printAffiliationsAndNotice{}

\begin{abstract} 

A major drawback of adversarially robust models, in particular for large scale datasets like ImageNet, is the extremely long training time compared to standard ones. Moreover, models should be robust not only to one $l_p$-threat model but ideally to all of them. In this paper we propose Extreme norm  Adversarial Training (E-AT) for multiple-norm robustness which is based on geometric properties of $l_p$-balls. E-AT costs up to three times less than other adversarial training methods for multiple-norm robustness. Using E-AT we show that for ImageNet a single epoch and for CIFAR-10 three epochs are sufficient to turn any $l_p$-robust model into a multiple-norm robust model.  In this way we get the first multiple-norm robust model for ImageNet and boost the state-of-the-art for multiple-norm robustness to more than $51\%$ on CIFAR-10.  Finally, we study the general transfer via fine-tuning of adversarial robustness between different individual $l_p$-threat models and 
improve the previous SOTA $l_1$-robustness on both CIFAR-10 and ImageNet. Extensive experiments show that our scheme works across datasets and architectures including vision transformers. 
\end{abstract}

\section{Introduction}
The problem of adversarial examples, that is small adversarial perturbations of the input \citep{SzeEtAl2014,KurGooBen2016a} changing the decision of a classifier, is a serious obstacle for the use of machine learning in safety-critical systems.
Many adversarial defenses have been proposed but most of them could  be broken  either by stronger attacks 
\citep{CarWag2016,AthEtAl2018,MosEtAl18} or using adaptive attacks \citep{tramer2020adaptive}. Apart from provable adversarial defenses which are however still restricted to rather simple CNNs \citep{WonEtAl18,GowEtAl18}, the only successful technique so far is adversarial training \citep{MadEtAl2018} and its improvements \citep{ZhaEtAl2019,CarEtAl19,wu2020adversarial,wu2021wider}. 
Current state-of-the-art results for $l_2$- and $l_\infty$-adversarial robustness 
on CIFAR-10 \cite{gowal2020uncovering,rebuffi2021data,gowal2021improving} exploit very large networks like WideResNet-70-16. One can anticipate that this trend of using evergrowing architectures will continue, as we can currently see for ImageNet \cite{dai2021coatnet,zhai2021scaling}. Therefore, especially considering that adversarial training is by itself more expensive than standard training, it is increasingly important that robust models can be quickly adapted to other tasks and threat models without the need for an excessively costly re-training from scratch. For this reason, in this paper we focus on fine-tuning which we believe will be an important topic in adversarial robustness in the future.

Moreover, while the community initially focused on adversarial examples for $l_\infty$-perturbations, there has recently been more interest in other $l_p$-attacks, such as $l_1$ and $l_2$, or perceptual threat models \citep{StutzCVPR2019, wong2020learning,laidlaw2021perceptual}. It is well known that robustness in one $l_p$-ball does not necessarily generalize to some other $l_q$-ball for $p\neq q$ \citep{kang2019transfer,TraBon2019}. However, in safety-critical systems we need robustness against all $l_p$-norms simultaneously which has triggered recent extensions of adversarial training for multiple $l_p$-norms \citep{TraBon2019,maini2020adversarial} and provable defenses for all $l_p$ with $p\geq 1$ \citep{croce2020provable}. 
\begin{figure*}[t] \centering
\includegraphics[width=2\columnwidth]{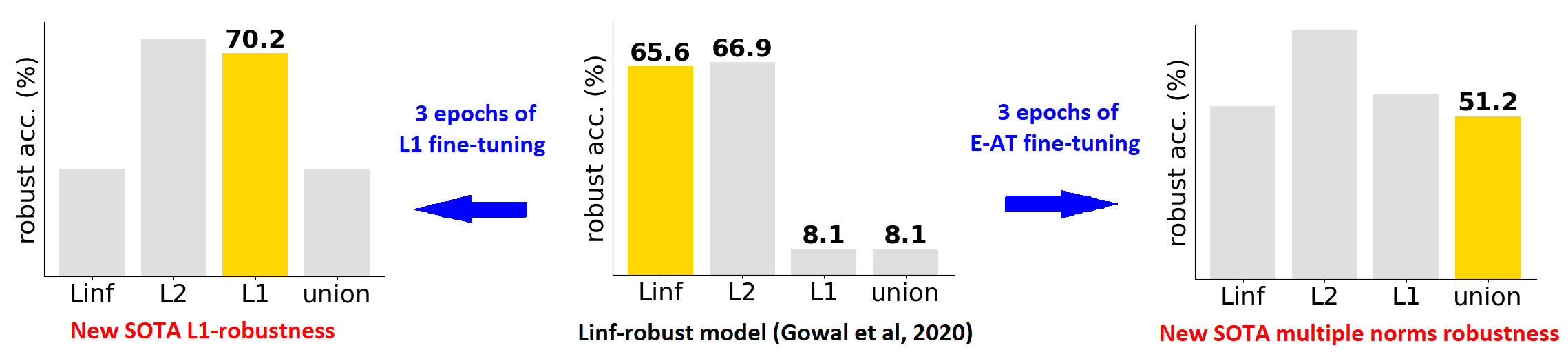}
\caption{We fine-tune for 3 epochs the WideResNet-70-16 on CIFAR-10 from \cite{gowal2020uncovering} with highest $l_\infty$-robustness to be either robust wrt $l_1$ (left) or with our E-AT to be robust wrt the union of the $l_\infty$-, $l_2$-, and $l_1$-threat models (right). We achieve state-of-the-art results in both threat models. The plots show the robust accuracy in the individual threat models and in their union for the initial $l_\infty$-robust classifier (middle) and the fine-tuned ones, with the target threat model highlighted.}\label{fig:teaser_ft}
\end{figure*}
In this paper we show that, using the geometry of the $l_p$-balls, the computationally expensive multiple norm training procedures of \citet{TraBon2019,maini2020adversarial}, which costs up to three times as much as normal adversarial training, can be replaced by a very effective and simple form of adaptively alternating between the two extreme norms, namely $l_1$ and $l_\infty$. This scheme achieves similar robustness for the union of the threat models to more costly previous approaches. Additionally, we show that on CIFAR-10 three epochs \MH{and on ImageNet even just a single epoch} of fine-tuning with our extreme norms adversarial training (E-AT) are sufficient to turn any $l_p$-robust model for $p \in \{1,2,\infty\}$ into a model which is robust against all $l_p$-threat models for $p \in \{1,2,\infty\}$, even if the original classifier was completely non-robust against one of 
them. 
Fine-tuning \hlrev{one of} the currently most robust networks in the $l_\infty$-threat model 
\cite{gowal2020uncovering} for CIFAR-10 we \hlrev{significantly} improve the current state-of-the-art performance for multiple norm robustness 
(robustness over the union of $l_1$, $l_2$ and $l_\infty$-balls)
\hlrev{without the need of training from scratch \MHREV{this large network} with existing expensive techniques}. Also, our E-AT fine-tuning scheme  yields \MHREV{in just one epoch} the first ImageNet model which is robust against multiple 
$l_p$-bounded attacks. 
Finally, we show that fine-tuning, again with only 
3 epochs for CIFAR-10 and 
one epoch for ImageNet, is sufficient to transfer robustness from one threat model to another one, which very quickly yields baselines for all threat models. In this way, starting from \MHREV{ $l_\infty$-robust classifiers} we achieve SOTA $l_1$-robust \MHREV{classifiers} on CIFAR-10 
\MHREV{and ImageNet}. These results are quite \MH{remarkable} as the original classifiers show no or 
very little $l_1$-robustness.
Fig.~\ref{fig:teaser_ft} summarizes the results of fine-tuning of robust classifiers on CIFAR-10 (see Table \ref{tab:CIFAR10-fine} and Table~\ref{tab:CIFAR10-fine-lp} for details), \MH{and Table \ref{tab:ImageNet-fine} for ImageNet shows that our results hold also for fine-tuning 
of transformer architectures.}

\section{Related Work}

\textbf{Adversarial training:}
Nowadays, adversarial training as formulated in \citet{MadEtAl2018} as a min-max optimization problem 
remains the only general method ensuring adversarial robustness across architectures and datasets \citep{AthEtAl2018}. Other types of  defenses use more sophisticated techniques, typically preventing the direct optimization of the attack. However, adaptive attacks specifically designed for these defenses have often shown that these alternative techniques are non-robust or much less robust than claimed \citep{tramer2020adaptive}. 
Recent improvements of adversarial training have been achieved by using different objectives \citep{ZhaEtAl2019}, unlabeled data \citep{CarEtAl19}, adversarial weights perturbations \citep{wu2020adversarial} and wider networks \cite{wu2021wider}. In \citet{gowal2020uncovering} several recent variants were systematically explored and for a very large architecture 
they obtain the most robust models for $l_\infty$ 
and $l_2$ 
for CIFAR-10, which we use for our fine-tuning experiments, \hlrev{only slightly improved later on by using particular data augmentation \citep{rebuffi2021data} or synthetic data \citep{gowal2021improving}. 
}

\textbf{Multiple-norm robustness:}
It was early on discovered that adversarial robustness against a specific $l_p$-threat model does typically not transfer to $l_q$-threat models for $p\neq q$ (see \citet{kang2019transfer,TraBon2019} for extensive studies). On the other hand to achieve really reliable machine learning models $l_p$-robustness wrt all $p$ is necessary.
The first approach to 
general robustness \cite{schott2018adversarially} uses multiple variational auto-encoders for an analysis by synthesis (ABS) architecture. While ABS is restricted to MNIST, it is robust against $l_0$, $l_2$ and $l_\infty$-attacks 
(although the $l_0$-robustness has been recently reduced with a stronger black-box attack \citep{croce2020sparsers}).
\citet{TraBon2019, maini2020adversarial, madaan2020learning} use variants of adversarial training to achieve robustness in multiple norms. Since these are the most similar methods to ours, we present them in detail below.
\citet{madaan2020learning} additionally proposes a meta-learning approach where one learns optimal noise to augment the samples and uses consistency regularization to enforce similar predictions on clean, augmented and adversarial samples.
Finally, \citet{StutzICML2020} combine adversarial training with a reject option: 
while that is effective, the comparison to normal adversarially trained models is difficult as their model is non-robust without the reject option.

\textbf{Provable robustness:} 
\citet{croce2020provable} motivated a regularization approach based on the geometry of the $l_p$-balls which enforces multiple-norm robustness during training: this allows to derive provable guarantees for multiple-norm robustness in contrast to the empirical evaluation of adversarial training. However, their approach works only for small network architectures and relatively small radii of the $l_p$-balls.

\textbf{Fine-tuning of robust models:}
Fine-tuning of an existing neural network is a commonly used technique in deep learning \citep{Goodfellow-et-al-2016} to quickly adapt a model to a different objective e.g. for language models \citep{howard2018universal}.
More recently, it has been shown in the area of adversarial robustness  that fine-tuning of pre-trained models, possibly using self-supervision, yields better adversarial robustness \citep{hendrycks2019using,chen2020adversarial,xu2021adversarial}. \citet{jeddi2020simple} show that fine-tuning of non-robust models with $10$ epochs can yield robust models with the caveat that their robustness evaluation is done using only a single run of PGD with $20$ steps. We are not aware of any prior work discussing fine-tuning to transform a robust model wrt
a single $l_p$ into one robust wrt multiple threat models or another $l_q$. 

\textbf{Evaluation of adversarial robustness:}
Many white-box attacks for $l_\infty$ \citep{MadEtAl2018,gowal2019}, $l_2$ \citep{MadEtAl2018,CarWag2016} and $l_1$ 
\citep{CheEtAl2018,ModEtAl19,rony2020augmented} have been proposed  as well as several black box attacks \citep{BreRauBet18,liu2018signsgd,cheng2018query,AlDujaili2019ThereAN,MeuEtAl2019,zhao2019design} for different threat models. It has recently been shown that AutoAttack \citep{croce2020reliable}, a parameter-free ensemble of the white-box attacks APGD for the cross-entropy and DLR-loss, 
FAB-attack \citep{CroHei2019} and the black-box Square Attack \citep{ACFH2019square} is reliably evaluating adversarial robustness for $l_2$ and $l_\infty$. AutoAttack has recently been extended to $l_1$ \citep{croce2021mind} 
outperforming all existing state-of-the-art attacks for $l_1$. 
\citet{croce2020reliable, croce2021mind} showed that on models defended with  adversarial training the two versions of APGD (with budget as in AutoAttack) already give an accurate robustness evaluation. As we have to evaluate very large models always for three threat models, we use those as a strong standard attack in our experiments.

\begin{table*}[h]
\caption{\textbf{CIFAR-10 - Other methods vs E-AT for fine-tuning:} We fine-tune with different methods for multiple norms for 3 epochs the RN-18 robust wrt $l_\infty$  (mean and standard deviation of the clean and robust accuracy over 5 seeds is reported).
We report clean performance, robust accuracy in each $l_p$-threat model, their average and the robust accuracy in their union (all values in percentage).
} \label{tab:ft_othermethods} \vspace{2mm}
\centering {\small 
\tabcolsep=2pt
\begin{tabular}{L{23mm} | *{5}{|C{20mm}} |>{\columncolor{LightCyan}}C{20mm} |>{\columncolor{LightCyan}}C{15mm}}
\textit{model} & \textit{clean} & $l_\infty$ ($\epsilon_\infty=\frac{8}{255}$)
&$l_2$ ($\epsilon_2=0.5$) & $l_1$ ($\epsilon_1=12$) & \textit{average} &\textit{union} & \textit{time/epoch} \\
\toprule
RN-18
$l_\infty$-AT & 83.7 & 48.1 & 59.8 & 7.7 & 38.5 & 7.7 & 151 s \\ 
\hfill + SAT & 83.5 $\pm$ 0.2 & 43.5 $\pm$ 0.2 & 68.0 $\pm$ 0.4 & 47.4 $\pm$ 0.5 & 53.0 $\pm$ 0.2 & 41.0 $\pm$ 0.3 & 161 s \\ 
\hfill + AVG & 84.2 $\pm$ 0.4 & 43.3 $\pm$ 0.4 & 68.4 $\pm$ 0.6 & 46.9 $\pm$ 0.6 & 52.9 $\pm$ 0.4 & 40.6 $\pm$ 0.4 & 479 s \\ 
\hfill + MAX & 82.2 $\pm$ 0.3 & 45.2 $\pm$ 0.4 & 67.0 $\pm$ 0.7 & 46.1 $\pm$ 0.4 & 52.8 $\pm$ 0.3 & 42.2 $\pm$ 0.6 & 466 s \\ 
\hfill + MSD & 82.2 $\pm$ 0.4 & 44.9 $\pm$ 0.3 & 67.1 $\pm$ 0.6 & 47.2 $\pm$ 0.6 & 53.0 $\pm$ 0.4 & 42.6 $\pm$ 0.2 & 306 s \\ 
\hfill + E-AT & 82.7 $\pm$ 0.4 & 44.3 $\pm$ 0.6 & 68.1 $\pm$ 0.5 & 48.7 $\pm$ 0.5 & 53.7 $\pm$ 0.3 & 42.2 $\pm$ 0.8 & 160 s
\\ \bottomrule
\end{tabular}} \end{table*}

\subsection{Adversarial training for the union of $l_1$-, $l_2$- and $l_\infty$-balls}\label{sec:Mult-rob} 
Let us denote by $f_\theta:\R^d\rightarrow \R^K$ the classifier parameterized by $\theta \in \R^n$, with input $x\in \R^d$ and $f_\theta(x) \in \R^K$ where $K$ is the number of classes of the task. Let further $\D=\{(x_i, y_i)\}_i$  be the training set, with $y_i$ the correct label of $x_i$, and $\mathcal{L}:\R^K \times \R^K \rightarrow \R$ a given loss function. The aim is to enforce adversarial robustness in all multiple $l_p$-balls simultaneously,
i.e., defining $B_p(\epsilon_p)=\{x\in\R^d: \norm{x}_p \leq \epsilon_p\}$, the threat model is the union of the individual $l_p$-balls,
i.e. $\Delta = B_1(\epsilon_1) \cup B_2(\epsilon) \cup B_\infty(\epsilon_\infty)$. $\Delta$ is a non convex set, since in practice the $\epsilon_p$ are chosen such that no $l_p$-ball contains any of the others. In adversarial training the worst case loss for input perturbations in the threat model, $ \max_{\delta \in \Delta} \mathcal{L}(f_\theta(x_i + \delta), y_i)$, is minimized. Efficiently maximizing the loss $\mathcal{L}$ in the union of threat models is non-trivial and different approaches 
have been proposed. 

\textbf{MAX:} \citet{TraBon2019} suggest to run  the three attacks for each $B_p(\epsilon_p)$ for $p \in \{1,2,\infty\}$ independently and then use 
the one which realizes the highest loss, that is
\[ 
\maxop\limits_{p \in \{1,2,\infty\}} \maxop_{\delta \in B_p(\epsilon_p)} \mathcal{L}(f_\theta(x_i + \delta), y_i).\]
This training optimizes directly the worst case in the union but comes at the price of being nearly three times as expensive as normal adversarial training wrt a single $l_p$-ball.

\textbf{AVG:} 
Additionally, \citet{TraBon2019} suggest to run  the three attacks 
independently but  replace the inner maximization problem with
\[ \sum_{p \in \{1,2,\infty\}} \maxop_{\delta \in B_p(\epsilon_p)} \mathcal{L}(f_\theta(x_i + \delta), y_i),\]
and thus averaging the updates of all $l_p$-balls with the motivation of not ``wasting'' the computed attacks, in particular when the attained loss values are rather similar and thus the max is ambiguous. 
This has similar cost to MAX.

\noindent\textbf{MSD:} 
\citet{maini2020adversarial} argue that the correct way to maximize the loss in the union is to test during the PGD attacks all the three steepest ascent updates corresponding to the three norms (sign of the gradient for $l_\infty$, normalized gradient for $l_2$ and a smoothed $l_1$-step by using the top-$k$ components in magnitude of the gradient) and then take the step which yields the highest loss. \citet{maini2020adversarial} report that MSD outperforms both AVG and MAX, also in terms of a more stable training. As all three updates (forward passes) are tested but only one backward pass is needed (gradient is the same) this costs roughly two times as much as normal adversarial training.

\textbf{SAT:} \citet{madaan2020learning} introduce Stochastic Adversarial Training (SAT) which randomly samples $p \in \{1, 2, \infty\}$ for each batch and performs PGD only for the corresponding $l_p$-norm. While SAT has the same cost as standard adversarial training, \citet{madaan2020learning} report that it does not perform very well.

\begin{figure*}[t]\centering
  \includegraphics[width=.40\columnwidth]{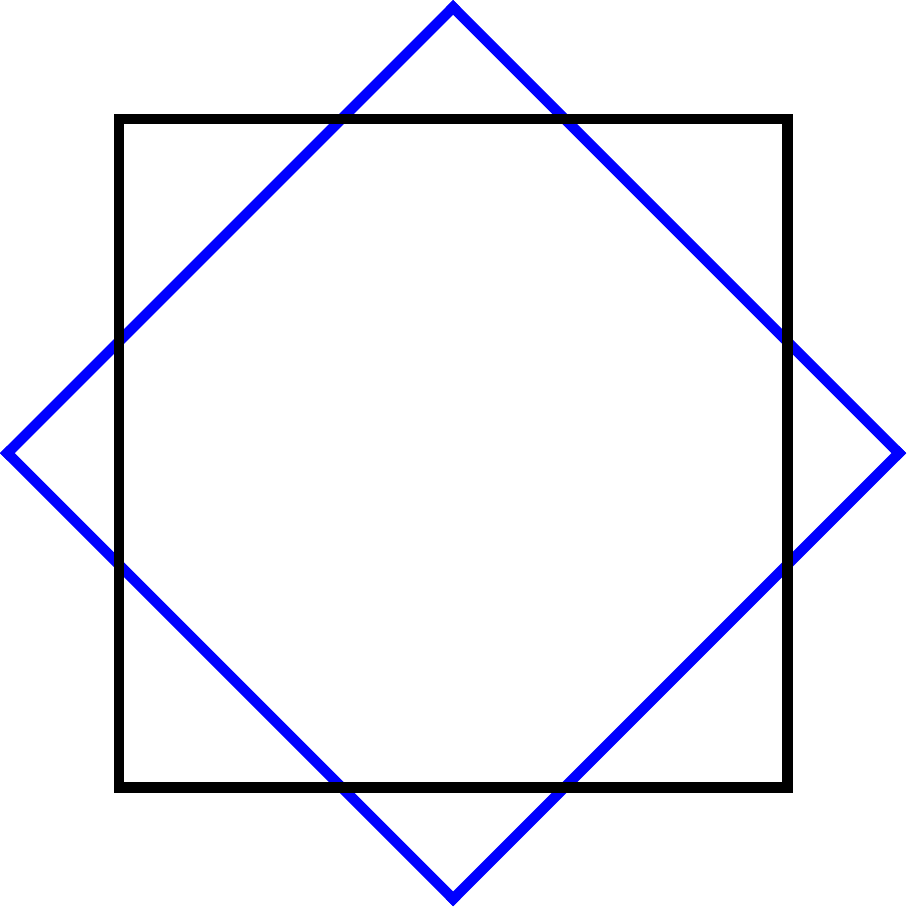}\hspace{4mm}
	\includegraphics[width=0.40\columnwidth]{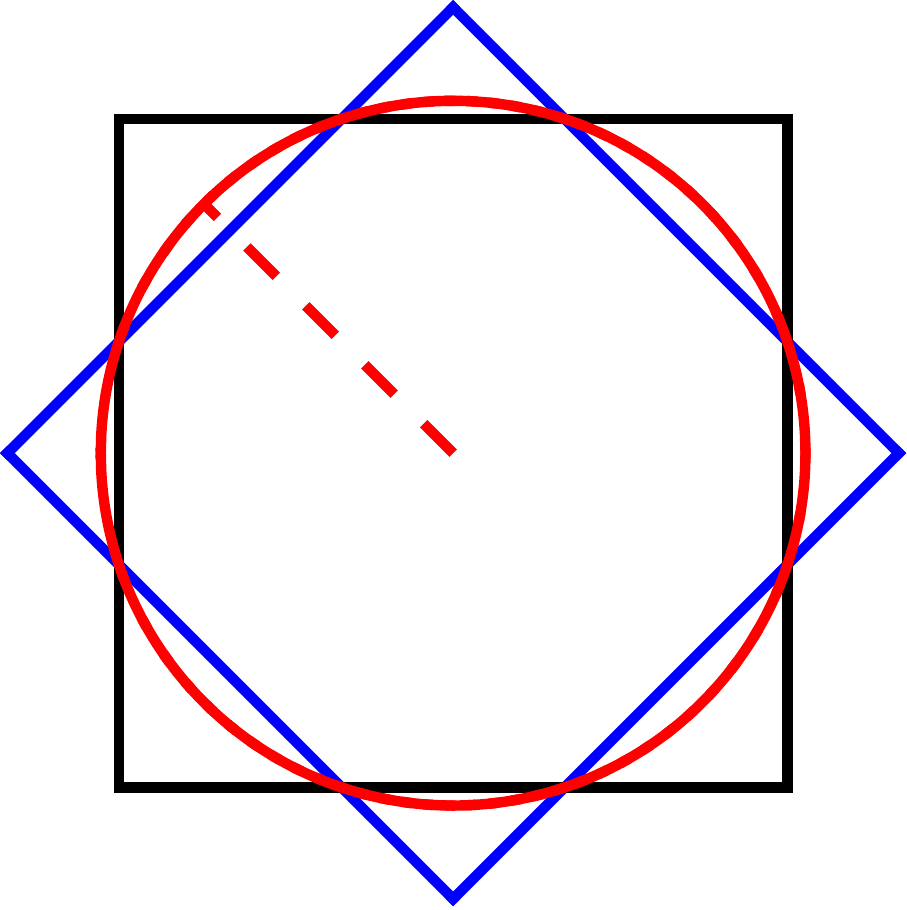}\hspace{4mm}
	\includegraphics[width=0.40\columnwidth]{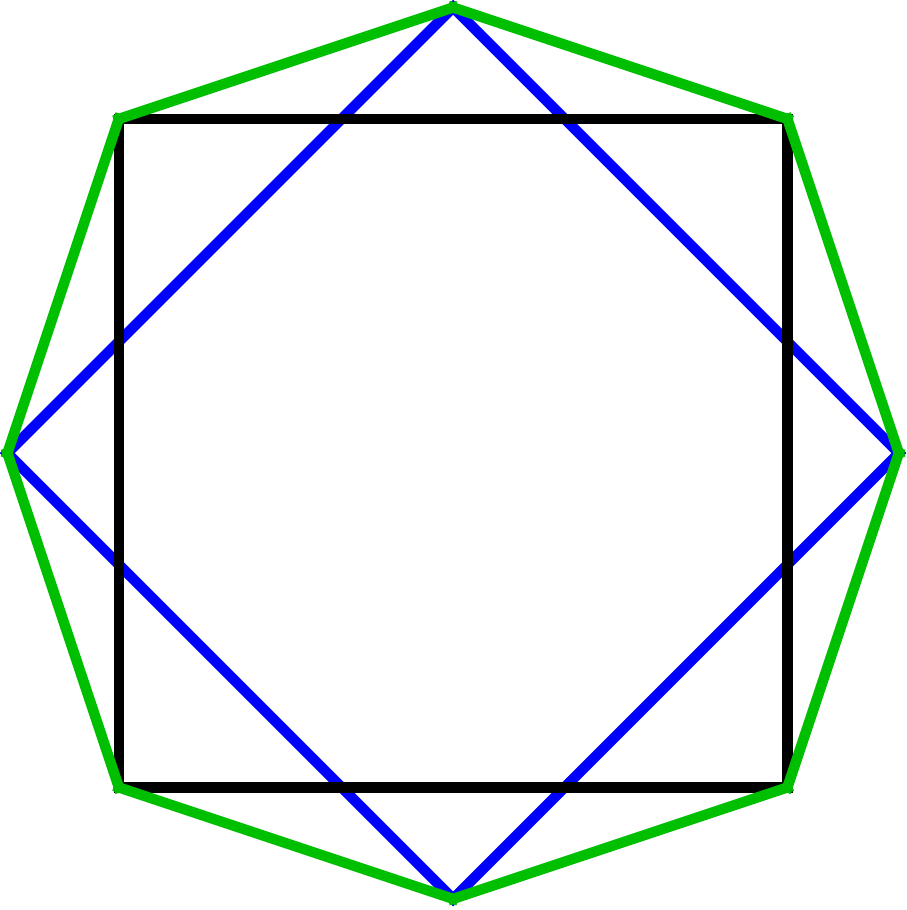}\hspace{4mm}
	\includegraphics[width=0.40\columnwidth]{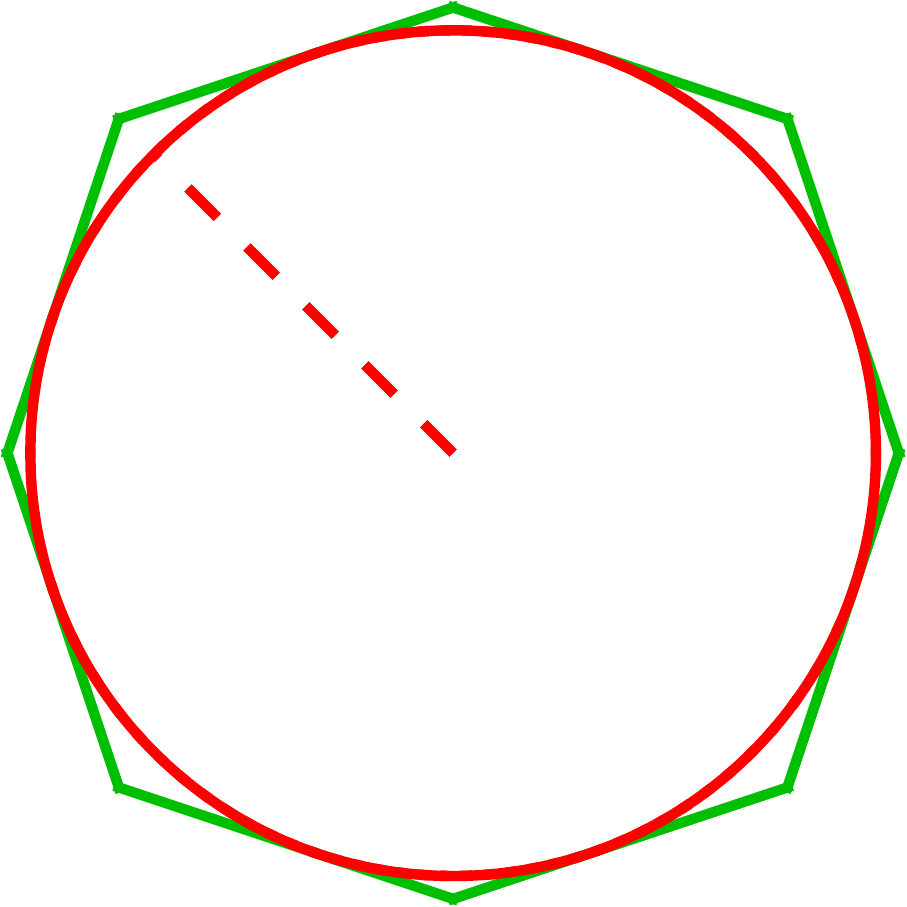} 
	\caption{Visualization of the $l_2$-ball contained in the union resp. the convex hull of the union of $l_1$- and $l_\infty$-balls in $\R^2$.
	\textbf{First:} co-centric $l_1$-ball (blue) and $l_\infty$-ball (black). \textbf{Second:} in red the largest $l_2$-ball 
	contained in the union of $l_1$- and $l_\infty$-ball. \textbf{Third:} in green the convex hull of the union of the $l_1$- and $l_\infty$-ball. \textbf{Fourth:} the largest $l_2$-ball (red) contained in the convex hull. The $l_2$-ball contained in the convex hull is significantly larger than that 
	in the union of $l_1$- and $l_\infty$-ball.}
	\label{fig:bounds_2}
\end{figure*}

\section{Fast Multiple-Norm Robustness via Extreme Norms Adversarial Training and Fine-Tuning}
\subsection{Multiple-norm robustness via fast fine-tuning of existing robust models}
Prior works \citep{TraBon2019, kang2019transfer} observed that models adversarially trained wrt $l_\infty$ give non trivial robustness to $l_2$-attacks, although lower than what one gets directly training against such attacks, and vice versa. This is confirmed by our evaluation 
in App.~\ref{sec:app_fulltraining}, where we also notice that $l_1$-AT provides good robust accuracy in $l_2$. On the other hand, training for $l_\infty$ resp. $l_1$ does not yield particular robustness to the dual norm, which is reasonable since the perturbations generated in the two threat models are very different, while the \MHREV{$l_2$-threat model is an intermedicate case which yields partial robustness against $l_\infty$ and $l_1$.}
Therefore, we propose to use models trained for robustness wrt a single norm as good initializations to achieve, within a small computational budget, multiple norms robustness. We test this by fine-tuning for 3 epochs an $l_\infty$-robust model on CIFAR-10 with the existing methods for multiple norms robustness and our E-AT. Table~\ref{tab:ft_othermethods} shows that a short fine-tuning (details in Sec.~\ref{sec:fine-tuning_large_models}) of a PreAct ResNet-18 \citep{he2016identity} trained with adversarial training wrt $l_\infty$ is effective in achieving competitive robustness in the union, not far from those of full training from random initialization (see Table~\ref{tab:fulltraining_rn18_singlecolumn}). However, the most effective methods, MAX and MSD, are 2-3x slower than standard adversarial training, while SAT is as fast as $l_\infty$-AT but performs slightly worse. E-AT aims at achieving the same results as MAX and MSD in the union while having complexity on par with SAT. 
Moreover, we report the average robustness in the 3 threat models, as done in prior works, where E-AT achieves the best results.
In App.~\ref{sec:app_ft_othermethods} we show that similar observations can be made when fine-tuning models initially $l_2$- or $l_1$-robust, 
\hlrev{and (Sec.~\ref{sec:fine-tuning_large_models}) 
are not specific to CIFAR-10 but generalize to ImageNet
.}

All previous methods assume that for achieving robustness to multiple norms each threat models has to be used at training time. In the following we first present an argument, using recent results from \citet{croce2020provable}, suggesting that this \MH{need} not be the case. Based on this analysis, we introduce our extreme norms adversarial training (E-AT) which achieves multiple norm robustness at the same price as training for a single $l_p$. 
Finally, we fine-tune with E-AT large robust models on CIFAR-10 and ImageNet.

\subsection{Geometry of the union of $l_p$-balls and their convex hull}
The main insight we use for E-AT is that a linear classifier which is robust in both an $l_1$- and an $l_\infty$-ball is also robust wrt the largest $l_p$-ball for $1\leq p \leq \infty$ which fits into the 
convex hull of the union of the $l_1$- and $l_\infty$-ball. This ball is significantly larger than the largest $l_p$-ball 
contained into the 
union of the $l_1$- and $l_\infty$-ball (see Fig.~\ref{fig:bounds_2}). Thus it is sufficient to be robust wrt the two ``extreme'' norms $l_1$ and $l_\infty$ to ensure robustness. 
While this is exact for affine classifiers, we conjecture that for neural networks this will at least hold approximately true 
(note that typical ReLU-networks yield piecewise affine classifiers \citep{AroEtAl2018}) and for the model it is the most efficient way in terms of capacity to be $l_1$- and $l_\infty$-robust.

We now state the main results from \citet{croce2020provable} which are the basis for our E-AT. We work in the non-trivial setting, where the balls are not included in each other 
as otherwise the problem of enforcing multiple norms (including $l_1$ or $l_\infty$) robustness 
boils down again to single norm robustness. 
For this, it has to hold  $\epsilon_1\in (\epsilon_\infty,d\epsilon_\infty)$: for CIFAR-10,  $\epsilon_\infty=\frac{8}{255}$ and dimension $d=3072$ yield an upper bound $\epsilon_1\leq 96.38$ which is far higher than $\epsilon_1=12$ commonly used for $l_1$-threat models. We denote by  $U_{1,\infty}(\epsilon_1,\epsilon_\infty)=B_1(\epsilon_1) \cup B_\infty(\epsilon_\infty)$ the union of the $l_1$- and $l_\infty$-balls. One can then ask the obvious question: how much $l_p$-robustness do I get for $1<p<\infty$ from a classifier robust in $U_{1,\infty}(\epsilon_1,\epsilon_\infty)$? 
\begin{proposition}[\citet{croce2020provable}] \label{prop:bound_2} If $d\geq 2$ and 
 $\epsilon_1\in (\epsilon_\infty,d\epsilon_\infty)$, then
	\begin{align}\minop_{x\in \R^d \setminus U_{1,\infty}(\epsilon_1,\epsilon_\infty)} \norm{x}_p = \left( \epsilon_\infty^p + \frac{(\epsilon_1-\epsilon_\infty)^p}{(d-1)^{p-1}}\right)^{\frac{1}{p}}.\label{eq:bounds_2} \end{align}
\end{proposition}
Thus a classifier which is robust for the union $U_{1,\infty}(\epsilon_1,\epsilon_\infty)$ has automatically 
a non-trivial robustness for all intermediate $l_p$-norms. For us the case $p=2$ is most interesting which
given that $\epsilon_1 \gg \epsilon_\infty$ can be tightly upper bounded as
\begin{align}
\epsilon_2:=\minop_{x\in \R^d \setminus U_{1,\infty}(\epsilon_1,\epsilon_\infty)} \norm{x}_2 \; \leq \; \sqrt{ \epsilon_\infty^2 + \frac{\epsilon_1^2}{d-1}}.\label{eq:l2-union} \end{align}
For CIFAR-10 
one gets $\epsilon_2\leq 0.2188$. As the radius of the $l_2$-threat model 
is usually chosen as $\epsilon_2=0.5$, robustness in the union alone would not be sufficient to achieve the desired robustness wrt $l_p$ for $p\in \{1,2,\infty\}$. 
However, if we consider affine classifiers then a guarantee for $B_1(\epsilon_1)$ \textit{and} $B_\infty(\epsilon_\infty)$ implies a guarantee with respect to the convex hull $C$ of their union $B_1(\epsilon_1) \cup B_\infty(\epsilon_\infty)$ as an affine classifier generates a half-space and thus only the extreme points of $B_1(\epsilon_1)$ resp. $B_\infty(\epsilon_\infty)$ matter 
(see Figure \ref{fig:bounds_2} for 
an illustration).
\begin{theorem}[\citet{croce2020provable}] Let $C$ be the convex hull of $B_1(\epsilon_1) \cup B_\infty(\epsilon_\infty)$. 
If $d\geq 2$ and $\epsilon_1\in (\epsilon_\infty,d\epsilon_\infty)$, then 
	\begin{equation} \minop_{x\in \R^d \setminus C} \norm{x}_p = \frac{\epsilon_1}{\left(\nicefrac{\epsilon_1}{\epsilon_\infty} -\alpha +\alpha^{q}\right)^{\nicefrac{1}{q}}},\label{eq:bounds_convex} \end{equation}
	where $\alpha = \frac{\epsilon_1}{\epsilon_\infty} - \lfloor \frac{\epsilon_1}{\epsilon_\infty}\rfloor$ and $\frac{1}{p}+\frac{1}{q}=1$. \label{th:bounds_convex}
\end{theorem}
As standard architectures using ReLU activation function yield a piecewise affine classifier one can anticipate that this result gives at least a rule of thumb on the expected $l_p$-robustness when one is $l_1$- and $l_\infty$-robust.
Again with the choice of $\epsilon_1,\epsilon_\infty$ from above one gets for the radius of the $l_2$-ball that fits into the convex hull $C$ of the union of $B_1(\epsilon_1)$ and $B_\infty(\epsilon_\infty)$:
\begin{equation} \epsilon_2:=\minop_{x\in \R^d \setminus C} \norm{x}_2 = \frac{\epsilon_1}{\sqrt{\nicefrac{\epsilon_1}{\epsilon_\infty} -\alpha +\alpha^{2}}}\approx 0.6178.\label{eq:bounds_convex_val} \end{equation}
Thus for a desired $l_2$-robustness with radius less than $0.6178$ it is sufficient for an affine classifier, and at least plausible for a ReLU network, to enforce $l_1$-robustness with $\epsilon_1=12$ and $l_\infty$-robustness with $\epsilon_\infty=\frac{8}{255}$. 
This 
motivates our extreme norms adversarial training (E-AT) and fine-tuning.

\subsection{Extreme norms adversarial training (E-AT)}
In light of the geometrical argument presented in the previous section, we propose to train only on adversarial perturbations for the $l_\infty$- and $l_1$-threat models if the $l_2$-radius obtained from Theorem \ref{th:bounds_convex} is larger than the radius $\epsilon_2$ of the $l_2$-threat model. In this case it is sufficient to just train for the extremes $l_1$ and $l_\infty$ in order to achieve robustness also to the intermediate $l_p$-attacks with $p\in(1, \infty)$. 
Since we seek a method as expensive as standard adversarial training, for each batch we do either the $l_1$- or the $l_\infty$-attack.
For full training from a random initialization simply alternating or sampling uniformly at random from the $l_1$- and $l_\infty$-attack works already well. However, for very quick fine-tuning, e.g. just one epoch in the case of ImageNet, for multiple norm robustness from an existing classifier robust wrt a single threat model, one has to take into account the existing robustness of the model. Thus we use an adaptive sampling strategy based on the running averages, reset at every epoch, of the robust training errors $\textrm{rerr}_1$ and $\textrm{rerr}_\infty$ (note that these running averages are computed just from averaging the robust error on the batches where the respective attack has been performed, thus no extra attacks are necessary), such that the probability for sampling the $l_p$-threat model is
\begin{equation}\label{eq:probl1linf}
\frac{\textrm{rerr}_p}{\textrm{rerr}_1+\textrm{rerr}_\infty}, \quad \textrm{for} \; p \in \{1,\infty\}.
\end{equation}
The motivation for this sampling scheme is that the robust error in the union $\Delta$ is mainly influenced by the worst threat model. We show in more details the effect of the biased sampling in E-AT fine-tuning in Sec.~\ref{sec:app_abl_seed_sampling}.

\begin{table*}[t] \centering \caption{\label{tab:CIFAR10-fine}\textbf{CIFAR-10 - 3 epochs of 
E-AT fine-tuning on $l_p$-robust models:} 
We fine-tune with E-AT models robust wrt a single $l_p$-norm, 
and report the robust accuracy on 1000 test points for all threat models and the difference 
to the initial classifier. 
(*) 
uses extra data.
} \vspace{2mm}
{\small
\tabcolsep=2.5pt
\begin{tabular}{C{28mm} L{37mm} | *{4}{| *{2}{C{8mm}}} 
|*{2}{>{\columncolor{LightCyan}}C{8mm}}}
&\textit{model} & \textit{clean} && \multicolumn{2}{|l}{$l_\infty$ ($\epsilon_\infty=\frac{8}{255}$)} &\multicolumn{2}{|l}{$l_2$ ($\epsilon_2=0.5$)} & \multicolumn{2}{|l|}{$l_1$ ($\epsilon_1=12$)} & 
\textit{union} & \\
\toprule
\multirow{10}{*}{\bfseries \makecell{Fine-tuning \\ $l_\infty$-robust models}} &
RN-50  - $l_\infty$ & 88.7 & & 50.9 & & 59.4 & & 5.0 & & 5.0 &\\ 
&
 \citep{robustness} \hfill + FT & 86.2 & \textcolor{red}{-2.5} & 46.0 & \textcolor{red}{-4.9} & 70.1 & \textcolor{blue}{10.7} & 49.2 & \textcolor{blue}{44.2} & 43.4 & \textcolor{blue}{38.4}\\ 
 
 \cmidrule{2-12} &
WRN-34-20  - $l_\infty$ & 87.2 & & 56.6 & & 63.7 & & 8.5 & & 8.5 &\\ 
&
 \citep{gowal2020uncovering} \hfill + FT & 88.3 & \textcolor{blue}{1.1} & 49.3 & \textcolor{red}{-7.3} & 71.8 & \textcolor{blue}{8.1} & 51.2 & \textcolor{blue}{42.7} & 46.2 & \textcolor{blue}{37.7}\\ 
 
 \cmidrule{2-12} &
WRN-28-10  - $l_\infty$ (*) & 90.3 & & 59.1 & & 65.7 & & 8.0 & & 8.0 &\\
&
 \citep{CarEtAl19} \hfill + FT & 90.3 & \textcolor{red}{0.0} & 52.6 & \textcolor{red}{-6.5} & 74.7 & \textcolor{blue}{9.0} & 54.0 & \textcolor{blue}{46.0} & 48.7 & \textcolor{blue}{40.7}\\ 
 
 \cmidrule{2-12}&
WRN-28-10  - $l_\infty$ (*) & 89.9 & & 62.9 & & 67.2 & & 10.8 & & 10.8 &\\ 
&
 \citep{gowal2020uncovering} \hfill + FT & 91.2 & \textcolor{blue}{1.3} & 53.9 & \textcolor{red}{-9.0} & 76.0 & \textcolor{blue}{8.8} & 56.9 & \textcolor{blue}{46.1} & 50.1 & \textcolor{blue}{39.3}\\ 
 
 \cmidrule{2-12} &
WRN-70-16  - $l_\infty$ (*) & 90.7 & & 65.6 & & 66.9 & & 8.1 & & 8.1 &\\
&
 \citep{gowal2020uncovering} \hfill + FT & 91.6 & \textcolor{blue}{0.9} & 54.3 & \textcolor{red}{-11.3} & 78.2 & \textcolor{blue}{11.3} & 58.3 & \textcolor{blue}{50.2} & 51.2 & \textcolor{blue}{43.1}\\ 
 \midrule
 \midrule
\multirow{6}{*}{\bfseries \makecell{Fine-tuning \\ $l_2$-robust models}}&
RN-50  - $l_2$ & 91.5 & & 29.7 & & 70.3 & & 27.0 & & 23.0 &\\ 
&
 \citep{robustness} \hfill + FT & 87.8 & \textcolor{red}{-3.7} & 43.1 & \textcolor{blue}{13.4} & 70.8 & \textcolor{blue}{0.5} & 50.2 & \textcolor{blue}{23.2} & 41.7 & \textcolor{blue}{18.7}\\ 
 
 \cmidrule{2-12} &
RN-50  - $l_2$ (*) & 91.1 & & 37.7 & & 73.4 & & 31.2 & & 28.8 &\\ 
&
 \citep{augustin2020} \hfill + FT & 87.0 & \textcolor{red}{-4.1} & 47.2 & \textcolor{blue}{9.5} & 70.4 & \textcolor{red}{-3.0} & 54.1 & \textcolor{blue}{22.9} & 46.0 & \textcolor{blue}{17.2}\\ 
 
 \cmidrule{2-12} &
WRN-70-16  - $l_2$ (*) & 94.1 & & 43.1 & & 81.7 & & 34.6 & & 32.4 &\\
&
 \citep{gowal2020uncovering} \hfill + FT & 91.2 & \textcolor{red}{-2.9} & 51.9 & \textcolor{blue}{8.8} & 79.2 & \textcolor{red}{-2.5} & 58.8 & \textcolor{blue}{24.2} & 49.7 & \textcolor{blue}{17.3}\\ 
 \midrule\midrule
\multirow{2}{*}{\bfseries \makecell{Fine-tuning \\ $l_1$-robust models}} &
RN-18  - $l_1$ & 87.1 & & 22.0 & & 64.8 & & 60.3 & & 22.0 &\\ 
&
 \citep{croce2021mind} \hfill + FT & 83.5 & \textcolor{red}{-3.6} & 40.3 & \textcolor{blue}{18.3} & 68.1 & \textcolor{blue}{3.3} & 55.7 & \textcolor{red}{-4.6} & 40.1 & \textcolor{blue}{18.1}
\\
\bottomrule
\end{tabular} }
\end{table*}

\subsection{
Scaling up multiple-norm robust models} \label{sec:fine-tuning_large_models}
The fact that multiple-norm robustness can be achieved via a short fine-tuning allows to use large architectures, which would be hard and expensive to train from random initialization in this more difficult threat model. Moreover, fine-tuning with E-AT has similar computational cost per epoch as standard adversarial training, \MH{and} since we aim at efficiency, we \MH{therefore} use it as main tool on large models.

\textbf{Experimental details:}
In the following we fine-tune with E-AT\footnote{Code available at \url{https://github.com/fra31/robust-finetuning}.} 
for 3 epochs on CIFAR-10 and 1 epoch on ImageNet-1k, starting with learning rate $0.05$ or $0.01$, depending on the model, and decreasing by a factor of $10$ every $1/3$ of \MH{the total number of finetuning epochs}. We do 10 steps of APGD in adversarial training for CIFAR-10, while 5 and 15 with $l_\infty$ and $l_1$ respectively on ImageNet as optimizing in the $l_1$-ball requires more iterations in that case. When the model was originally trained with extra data beyond the training set on CIFAR-10, we use the 500k images introduced by \citet{CarEtAl19} as additional data for fine-tuning (see also Sec.~\ref{sec:app_exps_details}).

\begin{table*} \centering \caption{\label{tab:ImageNet-fine}\textbf{ImageNet - Results of one epoch of E-AT fine-tuning of existing robust models:} We use existing models trained to be robust wrt a single $l_p$-ball (either $l_\infty$ or $l_2$) and fine-tune them for a single epoch \MH{for multiple-norm robustness} with our E-AT scheme.}
\vspace{2mm}
{\small
\tabcolsep=2.5pt
\begin{tabular}{C{29mm} L{37mm} | *{4}{| *{2}{C{8mm}}} |*{2}{>{\columncolor{LightCyan}}C{8mm}}}
&\textit{model} & \textit{clean} && \multicolumn{2}{|l}{$l_\infty$ ($\epsilon_\infty=\frac{4}{255}$)} &\multicolumn{2}{|l}{$l_2$ ($\epsilon_2=2$)} & \multicolumn{2}{|l|}{$l_1$ ($\epsilon_1=255$)} & 
union & \\
\toprule
\multirow{8}{*}{\bfseries \makecell{Fine-tuning \\ $l_\infty$-robust models}} &
RN-50  - $l_\infty$ & 62.9 & & 29.8 & & 17.7 & & 0.0 & & 0.0 &\\ 
& \citep{robustness} \hfill + FT & 58.0 & \textcolor{red}{-4.9} & 27.3 & \textcolor{red}{-2.5} & 41.1 & \textcolor{blue}{23.4} & 24.0 & \textcolor{blue}{24.0} & 21.7 & \textcolor{blue}{21.7}\\ 
 
 \cmidrule{2-12}
&
RN-50  - $l_\infty$ & 68.2 & & 36.7 & & 15.6 & & 0.0 & & 0.0 &\\ 
& \citep{bai2021are} \hfill + FT & 60.1 & \textcolor{red}{-8.1} & 29.2 & \textcolor{red}{-7.5} & 42.1 & \textcolor{blue}{26.5} & 24.5 & \textcolor{blue}{24.5} & 22.6 & \textcolor{blue}{22.6}\\ 
 
 \cmidrule{2-12}
&
DeiT-S  - $l_\infty$ & 66.4 & & 35.6 & & 40.1 & & 3.1 & & 3.1 &\\ 
&
\citep{bai2021are} \hfill + FT & 62.6 & \textcolor{red}{-3.8} & 32.2 & \textcolor{red}{-3.4} & 46.1 & \textcolor{blue}{6.0} & 24.8 & \textcolor{blue}{21.7} & 23.6 & \textcolor{blue}{20.5}\\
 
\cmidrule{2-12}
&XCiT-S - $l_\infty$ & 72.8 & & 41.7 & & 45.3 & & 2.7 & & 2.7 &\\ 
& \citep{debenedetti2022adversarially} \hfill + FT & 68.0 & \textcolor{red}{-4.8} & 36.4 & \textcolor{red}{-5.3} & 51.3 & \textcolor{blue}{6.0} & 28.4 & \textcolor{blue}{25.7} & 26.7 & \textcolor{blue}{24.0} \\
\midrule
 \midrule
\multirow{2}{*}{\bfseries \makecell{Fine-tuning \\ $l_2$-robust models}}&
RN-50  - $l_2$ & 58.7 & & 25.0 & & 40.5 & & 14.0 & & 13.5 &\\
&
 \citep{robustness} \hfill + FT & 56.7 & \textcolor{red}{-2.0} & 26.7 & \textcolor{blue}{1.7} & 41.0 & \textcolor{blue}{0.5} & 25.4 & \textcolor{blue}{11.4} & 23.1 & \textcolor{blue}{9.6}
\\ \bottomrule 
\end{tabular} }
\end{table*}

\begin{table}[h] \centering
\caption{\label{tab:ImageNet_ft_othermethods}\textbf{ImageNet - 1 epoch of fine-tuning of existing robust models:} We use existing models trained to be robust wrt a single $l_p$-ball (either $l_\infty$ or $l_2$) and fine-tune them for a single epoch \MH{for multiple-norm robustness} 
with SAT and E-AT.}
\vspace{2mm}
{\small
\tabcolsep=2.1pt
\begin{tabular}{L{22mm} | *{4}{| *{1}{C{10mm}}} |*{1}{>{\columncolor{LightCyan}}C{10mm}}}
\textit{model} & \textit{clean} & $l_\infty$ 
&$l_2$ 
& $l_1$ 
& 
\textit{union} \\
\toprule
RN-50 - $l_\infty$-AT
& 62.9 & 29.8 & 17.7 & 0.0 & 0.0 \\ 
\hfill + SAT & 59.4 & 26.5 & 38.8 & 21.1 & 19.4 \\ 
\hfill + E-AT & 58.0 & 27.3 & 41.1 & 24.0 & 21.7 \\ \midrule
RN-50 - $l_2$-AT
 & 58.7 & 25.0 & 40.5 & 14.0 & 13.5 \\ 
\hfill + SAT & 57.7 & 25.9 & 41.6 & 23.2 & 21.1 \\ 
\hfill + E-AT & 56.7 & 26.7 & 41.0 & 25.4 & 23.1
\\ \bottomrule
\end{tabular} } \end{table}

\textbf{CIFAR-10:} RobustBench \citep{croce2020robustbench} provides a collection of the currently most robust classifiers. We took \hlrev{a subset of} the most robust models, among those which do not use synthetic data, for $l_2$- and $l_\infty$-norm and the $l_1$-robust one from \citet{croce2021mind} (all are trained with the same radii $\epsilon_p$ as in our experiment). 
\hlrev{Note that we use the classifiers from \citet{gowal2020uncovering} (instead of those from \citet{rebuffi2021data}) since those were the best available ones at the time of the start of this project.}
We present in Table \ref{tab:CIFAR10-fine} the results. First of all the fine-tuning works for all tested architectures and results in many cases in stronger robustness in the union than for the specifically trained WideResNet-28-10 models (see Table~\ref{tab:training_rand_init_wrn2810_avg}). In particular, the most robust $l_\infty$-model from \citet{gowal2020uncovering} with $65.6\%$ $l_\infty$-robustness and only $8.1\%$ $l_1$-robustness can be fine-tuned to a multiple-norm robust model with $51.2\%$ robustness which is up to our knowledge the best reported multiple-norm robustness. \hlrev{While in general it is expected that larger architectures and extra data improve robustness (see e.g. RobustBench leaderboards), we could achieve such improvement without the high computational cost (and potential instabilities) of training large networks on an extended dataset from scratch.} Very interesting is that the $l_2$-robustness of $78.2\%$ is quite close to the $81.7\%$ $l_2$-robustness of the specifically $l_2$-trained model from \citet{gowal2020uncovering}. Moreover, the $l_1$-robustness of $58.3\%$ is close to the best reported one of $60.3\%$ \cite{croce2021mind} (however we improve this a lot in the next section) and the model has even higher clean accuracy. Clearly, this comes at the price of a significant loss in $l_\infty$ but this is to be expected. Striking is that fine-tuning the $l_2$-robust model from \citet{gowal2020uncovering} results in a very similar result. \hlrev{Finally, we observe that the $l_\infty$-threat model is the most challenging one, and fine-tuning \MHREV{$l_\infty$-}robust models yields the best robust accuracy in the union (when comparing models with the same architecture).}
In a nutshell, E-AT fine-tuning of existing $l_p$-robust models yields very efficient and competitive baselines for future research in this area.

\textbf{ImageNet:} 
We start with the 
$l_2$- resp. $l_\infty$-robust models from \citet{robustness}, \citet{bai2021are} and \citet{debenedetti2022adversarially} 
including the vision transformers DeiT small \citep{touvron2021training} and XCiT small \citep{el2021xcit} 
We use $\epsilon_2=2$ for the experiments as the robust accuracy is still in a reasonable range of $40\%$ and together with our choice of $\epsilon_1=255$ and the standard $\epsilon_\infty=\frac{4}{255}$ the $l_2$-radius from Theorem \ref{th:bounds_convex} is almost exactly $2$. 
The initial $l_\infty$-models are completely non-robust for $l_1$ but achieve, after fine-tuning, over $24\%$ $l_1$-robust accuracy and also the $l_2$-robust accuracy improves, 
at the price of a relatively small loss in $l_\infty$-robust and clean accuracy. Interestingly, the DeiT-S and XCiT-S models has already high robustness wrt $l_2$, unlike the RN-50s, which further improves thanks to E-AT, and the latter attains the best robustness in the union. For the $l_2$-robust model all robust accuracies improve as the original model was trained for $\epsilon=3$. 
Up to our knowledge no multiple-norm robustness has been reported before for ImageNet and thus these results are an important baseline. \hlrev{Finally, we show in Table~\ref{tab:ImageNet_ft_othermethods} that both SAT and E-AT are effective \MHREV{for fine-tuning} on ImageNet, and \MHREV{E-AT achieves the best robustness in the union} (we omit the other methods since they are computationally more expensive). We also observe that in this case $l_1$ is the most challenging threat model, and the best robustness in the union is achieved when fine-tuning the classifier (among those using RN-50 as architecture) trained wrt $l_2$ which already has non-trivial robustness wrt $l_1$. 
}

\textbf{Additional experiments:} \MH{Appendix}~\ref{sec:app_exps_ft}  contains further studies and details about \hlrev{fine-tuning with} E-AT, e.g. 
we \MHREV{report runtime and} show that fine-tuning a naturally trained model does not provide competitive robustness and leads to low clean accuracy.
Moreover, we show the stability of the scheme over random seeds, that increasing the number of epochs progressively improves the robustness in the union, and that even models trained to be robust wrt perceptual metrics can be used for E-AT fine-tuning. 

\subsection{\MH{Full training for} multiple norm robustness
from random initialization}
We evaluate the performance of the different 
methods for multiple-norm robustness when applied for full training from random initialization on CIFAR-10 (using the same $\epsilon_p$ as above). Table~\ref{tab:fulltraining_rn18_singlecolumn} reports the results, averaged over 3 runs, in every threat model: MAX and MSD attain the best robustness in the union, and E-AT is close to them and outperforms SAT. We recall that E-AT is 2-3x less expensive than MSD and MAX (see Table~\ref{tab:ft_othermethods}). We also include the performance of models trained \MHREV{to be robust} 
for single norms, which do not show high robustness in the union of the threat models. More details and further experiments with WideResNet-28-10 as architecture can be found in App.~\ref{sec:app_exps_details} and App.~\ref{sec:app_fulltraining}. 

\begin{table}[t]
\caption{\textbf{CIFAR-10 - Training from random initialization:} 
For full training of PreAct ResNet-18 with different methods for multiple norm robustness, we show robust accuracy in each $l_p$-threat model, their average, and the robustness in their union.
} \label{tab:fulltraining_rn18_singlecolumn} \vspace{2mm}
\centering {\small 
\tabcolsep=2pt
\begin{tabular}{L{15mm} | *{5}{|C{9mm}} |>{\columncolor{LightCyan}}C{9mm} 
}
\textit{model} & \textit{clean} & 
$l_\infty$ & $l_2$ & $l_1$ & \textit{avg.} &
\textit{union} 
\\
\toprule
$l_\infty$-AT & 84.0 & 48.1 & 59.7 & 6.3 & 38.0 & 6.3 \\ 
$l_2$-AT & 88.9 & 27.3 & 68.7 & 25.3 & 40.5 & 20.9 \\ 
$l_1$-AT & 85.9 & 22.1 & 64.9 & 59.5 & 48.8 & 22.1 \\
\midrule
SAT & 83.9 & 40.7 & 68.0 & 54.0 & 54.2 & 40.4 \\ 
AVG & 84.6 & 40.8 & 68.4 & 52.1 & 53.8 & 40.1 \\ 
MAX & 80.4 & 45.7 & 66.0 & 48.6 & 53.4 & 44.0 \\ 
MSD & 81.1 & 44.9 & 65.9 & 49.5 & 53.4 & 43.9 \\ 
E-AT & 81.9 & 43.0 & 66.4 & 53.0 & 54.2 & 42.4 
\\ \bottomrule
\end{tabular} } \end{table}

\begin{table*}[h]
\caption{\textbf{CIFAR-10 - Robustness against non $l_p$-bounded attacks:} We test the robustness of WRN-28-10 trained in different threat models against \MH{unseen} types of attacks. Moreover, we add the PAT model from \citet{laidlaw2021perceptual}, which uses RN-50 as architecture. 
}\label{tab:non_lp_attacks}
\vspace{2mm}
\centering {\small 
\tabcolsep=2pt
\begin{tabular}{L{11mm} | *{2}{|C{10mm}} | *{9}{|C{10mm}} |>{\columncolor{LightCyan}}C{10mm}
}
\textit{model} & \textit{clean} & 
\textit{comm. corr.}&$l_0$ & \textit{patches} & \textit{frames}& \textit{fog} & \textit{snow} &\textit{gabor} & \textit{elastic} & \textit{jpeg}
 &\textit{avg.}& \textit{union}  \\
\toprule
NAT & 94.4 & 71.6 & 0.1 & 8.1 & 2.6 & 47.3 & 3.9 & 35.0 & 0.2 & 0.0 & 12.2 & 0.0 \\ 
$l_\infty$-AT & 81.9 & 72.6 & 7.3 & 21.6 & 26.2 & 36.0 & 35.9 & 52.5 & 59.4 & 5.1 & 30.5 & 2.0 \\ 
$l_2$-AT & 87.8 & 79.2 & 13.2 & 25.0 & 17.7 & 44.9 & 22.1 & 43.5 & 56.6 & 14.0 & 29.6 & 4.5 \\ 
$l_1$-AT & 83.5 & 75.0 & 40.9 & 41.3 & 21.1 & 35.6 & 20.6 & 41.2 & 53.3 & 25.5 & 34.9 & 8.6 \\
PAT & 82.6 & 76.9 & 23.3 & 37.9 & 21.7 & 53.5 & 25.6 & 41.8 & 53.5 & 13.7 & 33.9 & 8.0 \\ \midrule
SAT & 80.5 & 72.0 & 38.7 & 36.7 & 29.3 & 33.5 & 29.0 & 49.8 & 57.0 & 37.4 & 38.9 & 13.8 \\ 
AVG & 82.0 & 73.6 & 39.7 & 36.8 & 30.8 & 37.2 & 21.1 & 49.9 & 58.1 & 30.4 & 38.0 & 10.9 \\
MAX & 80.1 & 71.3 & 35.1 & 34.6 & 32.7 & 34.5 & 35.0 & 53.4 & 58.5 & 33.5 & 39.7 & 15.3 \\ 
MSD & 81.0 & 71.7 & 36.9 & 35.0 & 31.8 & 34.6 & 26.4 & 51.5 & 59.7 & 33.4 & 38.7 & 12.9 \\
E-AT & 79.1 & 71.3 & 39.5 & 37.7 & 30.5 & 34.8 & 33.4 & 50.2 & 58.6 & 38.7 & 40.4 & 15.9
\\ \bottomrule
\end{tabular}} \end{table*}

\subsection{Robustness against unseen non $l_p$-bounded \MH{adversarial} attacks}
We \MH{investigate on CIFAR-10} to which extent  adversarial robustness achieved with multiple norms training generalizes to unseen and possibly very different threat models. We select three sparse attacks ($l_0$-bounded, patches and frames) and five adversarial corruptions (fog, snow, Gabor noise, elastic, $l_\infty$-JPEG) from \citet{kang2019testing}. Additionally, we compute the accuracy of the classifiers on the common corruptions, i.e. not adversarially optimized, of CIFAR-10-C \citep{HenDie2019}. Table~\ref{tab:non_lp_attacks} shows the results against such attacks of WRN-28-10 adversarially trained either wrt single norms  
\hlrev{or for multiple norm robustness. Additionally, we include} the PAT model from \citet{laidlaw2021perceptual}, which uses a RN-50 as architecture, \hlrev{is obtained with perceptual adversarial training,} \MH{and has been shown to be robust to unseen attacks}, and a \MH{normally} trained model \hlrev{(NAT)} as baseline. The models trained wrt multiple norms show much higher robustness in both the union \MHREV{(worst case robustness over all threat models, \hlrev{excluding common corruptions})} and average \MHREV{robustness against} these attacks: \hlrev{in particular, MAX and E-AT achieve} almost \MHREV{twice as high robustness} in the union than the best of the \hlrev{models not trained for multiple-norm robustness}, with only little loss in clean accuracy compared to the other robust classifiers. Moreover, for almost all threat models, 
\hlrev{these classifiers} attain the highest robustness or are close to it, while the best among the single $l_p$-robust models varies. \MH{Therefore training} for multiple norm robustness allows 
\MH{to generalize to some extent beyond the threat model used during training.}
Finally, E-AT outperforms even the PAT model which is trained wrt LPIPS and aims at generalization to unseen attacks. More details in App.~\ref{sec:app_otherattacks}.

\subsection{Effect of varying the radii of the $l_\infty$- and $l_1$-ball on the robustness wrt $l_2$} \label{sec:app_varying_radii}
\begin{figure}[t] \centering
\includegraphics[trim=0mm 8mm 0mm 0mm, width=0.75\columnwidth]{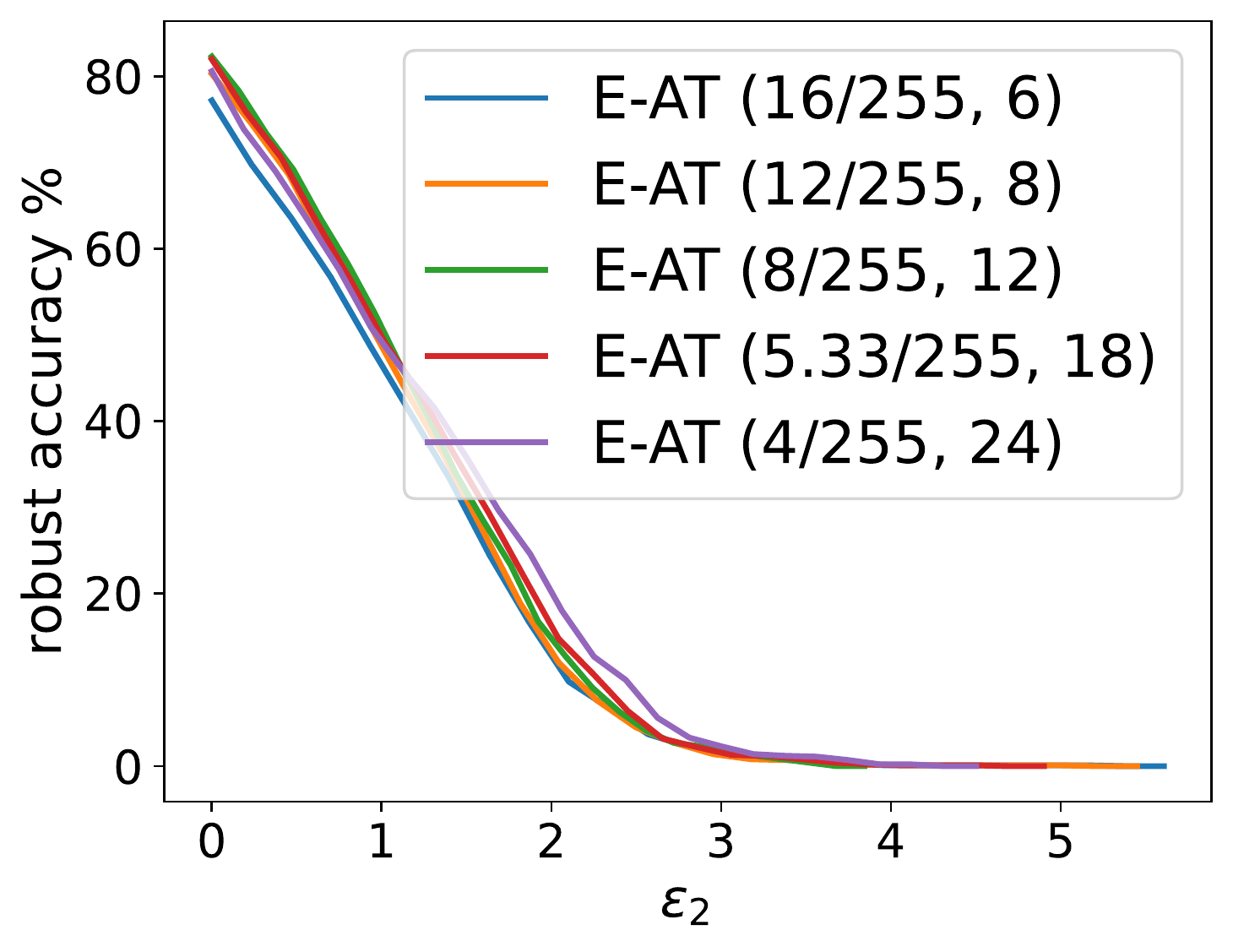}
\caption{\MHREV{$l_2$-robustness curves} of models \hlrev{obtained fine-tuning an $l_\infty$-robust RN-18} with E-AT on CIFAR-10 with different combinations  \MHREV{($\epsilon_\infty, \epsilon_1$)} of radii of the $l_\infty$- and $l_1$-ball.}\label{fig:curves_varying_radii}
\end{figure}
\hlrev{We study the effect of varying the radii of the $l_\infty$- and $l_1$-balls in E-AT on the robustness wrt $l_2$ of the resulting classifier. We fine-tune with E-AT the $l_\infty$-robust RN-18 (trained with $\epsilon_\infty=8/255$
) for 3 epochs with different pairs $(\epsilon_\infty, \epsilon_1)$ such that $\sqrt{\epsilon_\infty \cdot \epsilon_1}$ is constant at $\approx 0.62$, i.e. inducing, ignoring $\alpha$, the same $\epsilon_2$ according to Eq.~\eqref{eq:bounds_convex_val} (convex hull of the union), but very different  $\epsilon_2$ for Eq.~\eqref{eq:l2-union}  (union only), ranging from 0.13 (16/255, 6) to 0.44 (4/255, 24). 
Fig.~\ref{fig:curves_varying_radii} shows \MHREV{the $l_2$-robustness curves computed with FAB \citep{CroHei2019}.} \MHREV{For all combinations the $l_2$-robustness curves are similar}, even for the smaller $\epsilon_1$ (the small drops are related to the lower clean accuracy). \MHREV{Thus the statement of 
Theorem~\ref{th:bounds_convex} (convex hull) formulated for linear classifiers holds \MHREV{empirically} also for deep networks.}}

\begin{table*}[h]
\caption{\label{tab:CIFAR10-fine-lp}\textbf{Fine-tuning $l_p$-robust models to another threat model:} For each norm we fine-tune the most robust models wrt the other ones for 3 epochs for CIFAR-10 and 1 epoch for ImageNet and report clean and robust accuracy for all threat models. Even for the threat models where the robustness of the original model is low, the fine-tuning is sufficient to yield robustness almost at the same level of the specialized models with same architecture. For each threat model (column) we highlight in blue the model trained for the specific norm, in orange those only fine-tuned in the target norm. The values of the thresholds $\epsilon$ are the same used the multiple norms experiments.} \vspace{2mm}
\centering {\small
\tabcolsep=4pt
\centering \begin{tabular}{
cc}
\begin{tabular}[t]{L{25mm} || C{10mm}
*{3}{|*{1}{C{10mm}}} }
\multicolumn{5}{c}{\textbf{CIFAR-10}} \\
& \multirow{1}{*}{\textit{clean}} & $l_\infty$ & $l_2$ & $l_1$\\
\toprule
\multicolumn{5}{l}{WRN-70-16 \citep{gowal2020uncovering} - $l_\infty$ (*)}\\ \toprule
original & 90.7  & \cellcolor{blue!10}65.6  & 66.9  & 8.1 \\
 \hfill + FT wrt $l_2$ & 92.8 & 47.4 & \cellcolor{orange!10}80.0 & 34.0\\
 \hfill + FT wrt $l_1$ & 92.4 & 33.9 & 74.7 & \cellcolor{orange!10} \textbf{70.2}\\
 \midrule
\multicolumn{5}{l}{WRN-70-16 \citep{gowal2020uncovering} - $l_2$ (*)} \\
\toprule
original & 94.1  & 43.1  & \cellcolor{blue!10} 81.7  & 34.6 \\
 \hfill + FT wrt $l_\infty$ & 92.3 & \cellcolor{orange!10}58.5 & 73.5 & 11.4\\
 \hfill + FT wrt $l_1$ & 92.8 & 29.2 & 75.7 & \cellcolor{orange!10}68.9\\
 \midrule
\multicolumn{5}{l}{RN-18 \citep{croce2021mind} - $l_1$} \\
\toprule
original & 87.1  & 22.0  & 64.8 & \cellcolor{blue!10}60.3 \\
 \hfill + FT wrt $l_\infty$ & 82.7 & \cellcolor{orange!10}44.2 & 66.6 & 25.4\\
 \hfill + FT wrt $l_2$ & 88.0 & 31.0 & \cellcolor{orange!10}69.8 & 39.7
\\\bottomrule
\end{tabular} &
\begin{tabular}[t]{L{25mm} || C{10mm}
*{3}{|*{1}{C{10mm}}} }
\multicolumn{5}{c}{\textbf{ImageNet}} \\
& \multirow{1}{*}{\textit{clean}} & $l_\infty$ & $l_2$ & $l_1$\\
\toprule
\multicolumn{5}{l}{DeiT-S \citep{bai2021are} - $l_\infty$} \\ \toprule
original & 66.4  & \cellcolor{blue!10}35.6  & 40.1  & 3.1 \\ 
 \hfill + FT wrt $l_2$ & 66.5 & 31.2 & \cellcolor{orange!10} 46.1 & 9.6\\ 
 \hfill + FT wrt $l_1$ & 61.0 & 23.9 & 42.9 & \cellcolor{orange!10} 
 30.1
 \\ \midrule
\multicolumn{5}{l}{XCiT-S \citep{debenedetti2022adversarially} - $l_\infty$} \\ \toprule
original & 72.8  & \cellcolor{blue!10}41.7  & 45.3  & 2.7 \\ 
 \hfill + FT wrt $l_2$ & 71.5 & 35.9 & \cellcolor{orange!10}51.4 & 9.5\\ 
 \hfill + FT wrt $l_1$ & 65.8 & 25.2 & 47.1 & \cellcolor{orange!10}\textbf{33.9}\\
 \midrule
\multicolumn{5}{l}{RN-50 \citep{robustness} - $l_2$} \\ \toprule
original & 58.7  & 25.0  & \cellcolor{blue!10}40.5  & 14.0 \\
 \hfill + FT wrt $l_\infty$ & 59.1 & \cellcolor{orange!10}31.5 & 40.1 & 7.5\\
\hfill + FT wrt $l_1$ & 56.8 & 18.0 & 37.1 & \cellcolor{orange!10}28.7\\
\bottomrule \end{tabular}
\end{tabular}}
\end{table*}

\section{Fine-Tuning $l_p$-Robust Models to Become $l_q$-Robust for $p\neq q$} \label{sec:single_norm}
Motivated by our 
results for the multiple-norm threat model 
we study to which extent we can fine-tune an $l_p$-robust model to a $l_q$-robust model with $p \neq q$. Again the emphasis is on an extremely short fine-tuning time so that this is much faster than full adversarial training.

\textbf{CIFAR-10:} 
We fine-tune for 3 epochs the most $l_\infty$-robust model at $\epsilon_\infty=\frac{8}{255}$ of \citet{gowal2020uncovering} with adversarial training wrt $l_2$ and $l_1$ 
with $\epsilon_2=0.5$ and $\epsilon_1=12$. 
Table \ref{tab:CIFAR10-fine-lp} shows that fine-tuning for $l_1$-robustness yields $70.2\%$ $l_1$-robust accuracy which is $9.9\%$ more than the previously most robust model \hlrev{with a \MHREV{small} computational cost (\MHREV{to be fair} with a larger architecture and using extra data)}. Also we get a strikingly high $l_2$-robust accuracy for the $l_2$-fine-tuned model of $80.0\%$ not far away from the $81.7\%$ which ones gets by training for $l_2$ from scratch. Surprisingly, fine-tuning the $l_2$-robust model of \citet{gowal2020uncovering} wrt $l_1$ does not outperform the $l_1$-robustness achieved by fine-tuning their $l_\infty$-robust model. Interestingly, fine-tuning the $l_1$-robust PreAct ResNet-18 for $l_2$ yields a better $l_2$-robustness than $l_2$-training from scratch, see Table \ref{tab:training_rand_init_wrn2810_avg}. This shows again that fine-tuning of existing models \MH{requires} minimal effort and already provides strong baselines for adversarial robustness obtained by adversarial training from scratch and in some cases even outperforms them.

\textbf{ImageNet:} We fine-tune the $l_2$-
RN-50 
and the $l_\infty$-robust transformers DeiT-S and XCiT-S  used in the previous section to the other threat model respectively and wrt $l_1$. The results are in Table \ref{tab:CIFAR10-fine-lp} (those for a RN-50 robust wrt $l_\infty$ in App.~\ref{sec:app_indiv_threat_models}). 
For all models it is possible to achieve within one epoch of fine-tuning a non-trivial robustness in every threat model. 
In particular, note that those \MH{originally} trained for the $l_\infty$-threat model have very low 
robustness wrt $l_1$, while after fine-tuning XCiT-S 
achieves $33.9\%$ robust accuracy, even higher than what is obtained 
from the $l_2$-robust RN-50 
i.e. $28.7\%$. 
Note that compared to fine-tuning for multiple-norm robustness (see Table~\ref{tab:ImageNet-fine}), fine-tuning 
XCiT-S specifically for $l_2$ yields almost the same robust accuracy but $3.5\%$ better clean performance, while, \hlrev{when fine-tuning it} for $l_1$, the $l_1$-robust accuracy is 
$5.5\%$ \hlrev{higher}. Up to our knowledge our ImageNet models are the first ones for which $l_1$-robustness is reported.

\section{Conclusion}
Based on the geometry of the $l_p$-balls we have introduced E-AT, a novel training scheme for multiple-norm robustness which achieves comparable adversarial robustness in the union while being significantly faster. We also show for the first time that fine-tuning can be used to transfer adversarial robustness from a single $l_p$-threat model to the multiple norms one, and that one can even 
obtain an $l_q$-robust classifier with a quick fine-tuning of an $l_p$-robust one with $p\neq q$. This yields strong baselines for future research. We have in this way generated models with SOTA performance for multiple-norm and $l_1$-robustness on CIFAR-10 and the first models on ImageNet 
with significant multiple-norm as well as $l_1$-robustness. \MH{This shows that fine-tuning is an excellent technique to avoid the increasingly high costs of training large adversarially robust models from scratch.}

\section*{Acknowledgements}
We acknowledge support from the German Federal Ministry of
Education and Research (BMBF) through the Tübingen AI Center (FKZ: 01IS18039A), the DFG Cluster of Excellence “Machine Learning – New
Perspectives for Science”, EXC 2064/1, project number 390727645, and by DFG grant 389792660 as part of TRR 248.

\bibliographystyle{icml2022}
%\bibliography{Literatur}

\clearpage
\appendix

\section{Experimental Details}\label{sec:app_exps_details}
For the comparison of training schemes we use for multiple-norm robustness we train PreAct ResNet-18 \citep{he2016identity} with softplus activation function for 80 epochs with initial learning rate of $0.05$ reduced by a factor of $10$ after 70 epochs. 
When training WideResNet-28-10 \citep{ZagKom2016} we use a cyclic schedule for the learning rate with maximum value $0.1$ for 30 epochs. We use SGD optimizer with momentum of $0.9$ and weight decay of $5 \cdot 10^{-4}$, batch size of 128. We use random cropping and horizontal flipping as augmentation techniques. For adversarial training of models robust wrt a single norm and with SAT and our novel scheme E-AT we use APGD with default parameters, while for the retrained AVG, MAX, and MSD we use PGD for $l_\infty$ (step size $\epsilon_\infty/4$) and $l_2$ (step size $\epsilon / 3$), SLIDE \citep{TraBon2019} for $l_1$ (standard parameters). For all methods we use 10 steps for the inner maximization problem in adversarial training (note that AVG and MAX repeat the attack for all threat models, and MSD tests multiple steps, thus they are more expensive). For all schemes we select the best performing checkpoint for the comparison \hlrev{(that is the one with the highest robustness)} when using the piecewise schedule, the final checkpoint with the cyclic schedule. Moreover, we use the TRADES-XENT loss (TRADES loss \citep{ZhaEtAl2019} with adversarial points maximizing the cross-entropy loss) since \citet{gowal2020uncovering} show that this gives slightly better robustness on CIFAR-10 without additional data, while we use standard adversarial training \citep{MadEtAl2018} for PreAct ResNet-18. We train the classifiers of MNG-AC \citep{madaan2020learning} with the original code, where we set $\epsilon_1=12$ and rescale the step size linearly. Finally, for the runtime comparison we run each method on a single Tesla V100 GPU.

For fine-tuning on CIFAR-10 we use 3 epochs and the same setup as for full training except for the learning rate schedule, since in this case we use as initial value the best performing one in $\{0.01, 0.05\}$ (the larger value works best for the smaller networks) and reduce it by a factor of $10$ at the beginning of each epoch. When the model was originally trained with extra data beyond the training set on CIFAR-10, we use the 500k images introduced by \cite{CarEtAl19} as additional data for fine-tuning, and each batch is split equally between standard and extra images, and we count 1 epoch when the whole standard training set has been used: note that in this way, using only 3 epochs not the whole pseudo-labelled dataset is exploited.

For fine-tuning on ImageNet we use 1 epoch, initial learning rate of $0.01$ ($0.0005$ for XCiT-S), reduced by a factor of $10$ every $1/3$ of training steps. We follow the setup of \cite{robustness} for data augmentation and setting batch size to 256 (except for the RN-50 from \citet{bai2021are} and XCiT-S for which we use 192 to fit into the GPU memory) and weight decay to $10^{-4}$. For adversarial training we use APGD with 5 steps for $l_\infty$ and $l_2$, 15 steps for $l_1$ since optimizing in the $l_1$-ball intersected with the box constraints is more challenging, see \cite{croce2021mind}.

\section{Additional Analysis, Evaluation and Experiments for 
E-AT}\label{sec:app_exps_random_init}
We here analyze in more details our E-AT scheme and expand the comparison to existing methods presented above.

\subsection{Robustness wrt $l_2$ of E-AT}
\begin{figure}[t] \centering {\small
\textbf{Robustness to $l_2$-attacks}}\\
\includegraphics[trim=0mm 8mm 0mm 0mm, width=0.75\columnwidth]{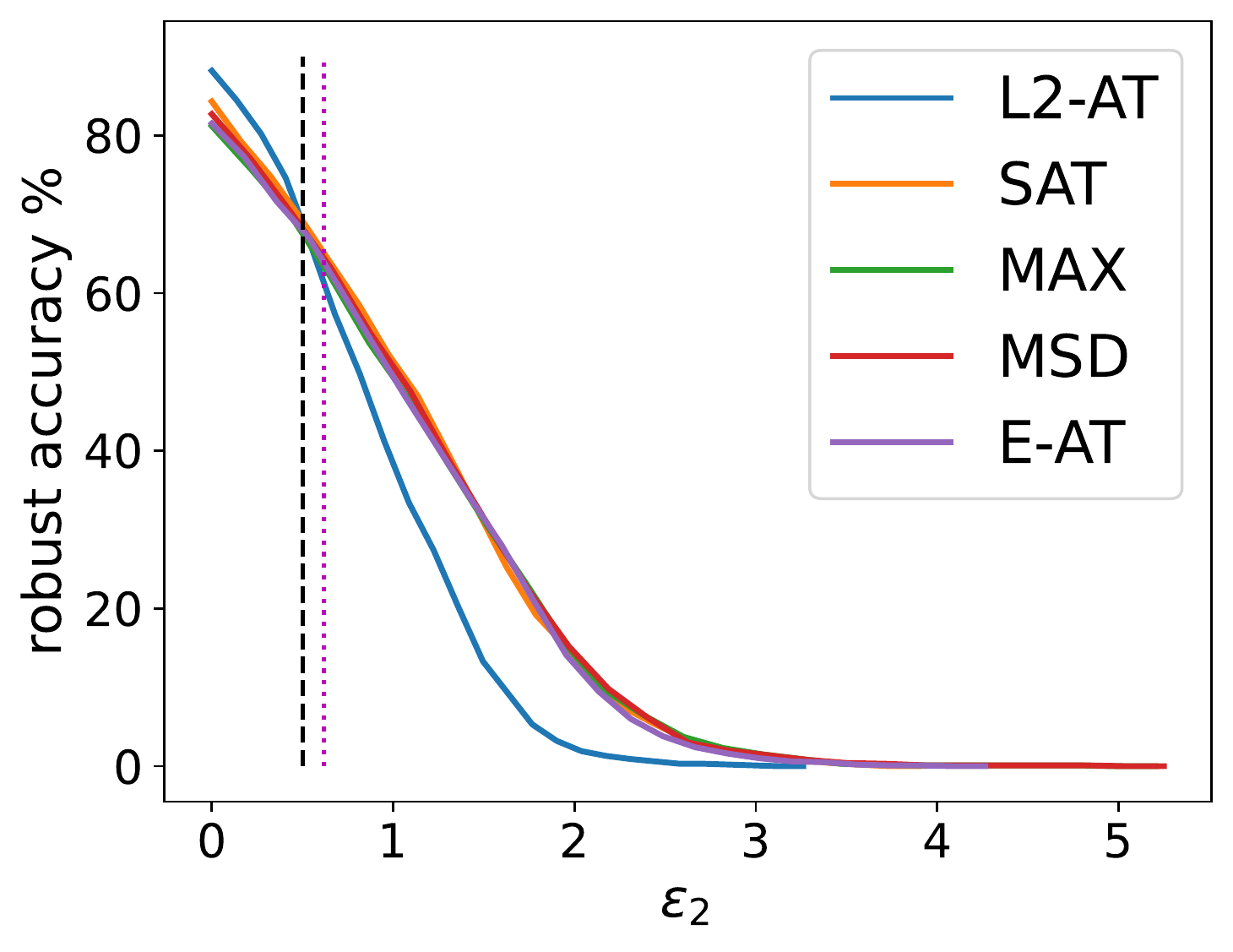}
\caption{\hlrev{$l_2$-robustness curves of a model trained with $l_2$-adversarial training (AT) for $\epsilon_2=0.5$ 
and methods for multiple norm robustness on CIFAR-10. Our E-AT}
is expected to yield robustness at $\epsilon_2=0.62$. \textbf{Although $l_2$-attacks are not used for training, our 
extreme-adv. training scheme E-AT yields $l_2$-robustness similar \hlrev{to the other methods and even} to the one obtained with specific $l_2$-adversarial training.}}
\label{fig:curves}\end{figure}

To show the effect of E-AT on $l_2$-robustness we plot in Fig.~\ref{fig:curves} the robust accuracy wrt $l_2$ computed with FAB \citep{CroHei2019}, which minimizes the size of the perturbations, as a function of the threshold $\epsilon_2$ for a PreAct ResNet-18 trained with \hlrev{either $l_2$-AT at $\epsilon_2=0.5$ or 
methods for multiple norm robustness} (see complete results for such models in Sec.~\ref{sec:app_fulltraining}). Theorem~\ref{th:bounds_convex} suggests that the extreme norms training provides robustness at $\epsilon_2 \approx 0.62$, which is confirmed by the plots. Although no $l_2$-attack has been used during training by the E-AT model, it has robustness wrt $l_2$ similar to that \hlrev{both of the techniques for multiple norms which use $l_2$-perturbations at training time and} of the classifier specifically trained for such threat model.

\subsection{Confirmation that 
adversarial training with the extreme norms is sufficient: Analysis of MAX and MSD training}
\begin{figure*}[h] \centering 
\tabcolsep=1pt
\includegraphics[width=0.70\columnwidth]{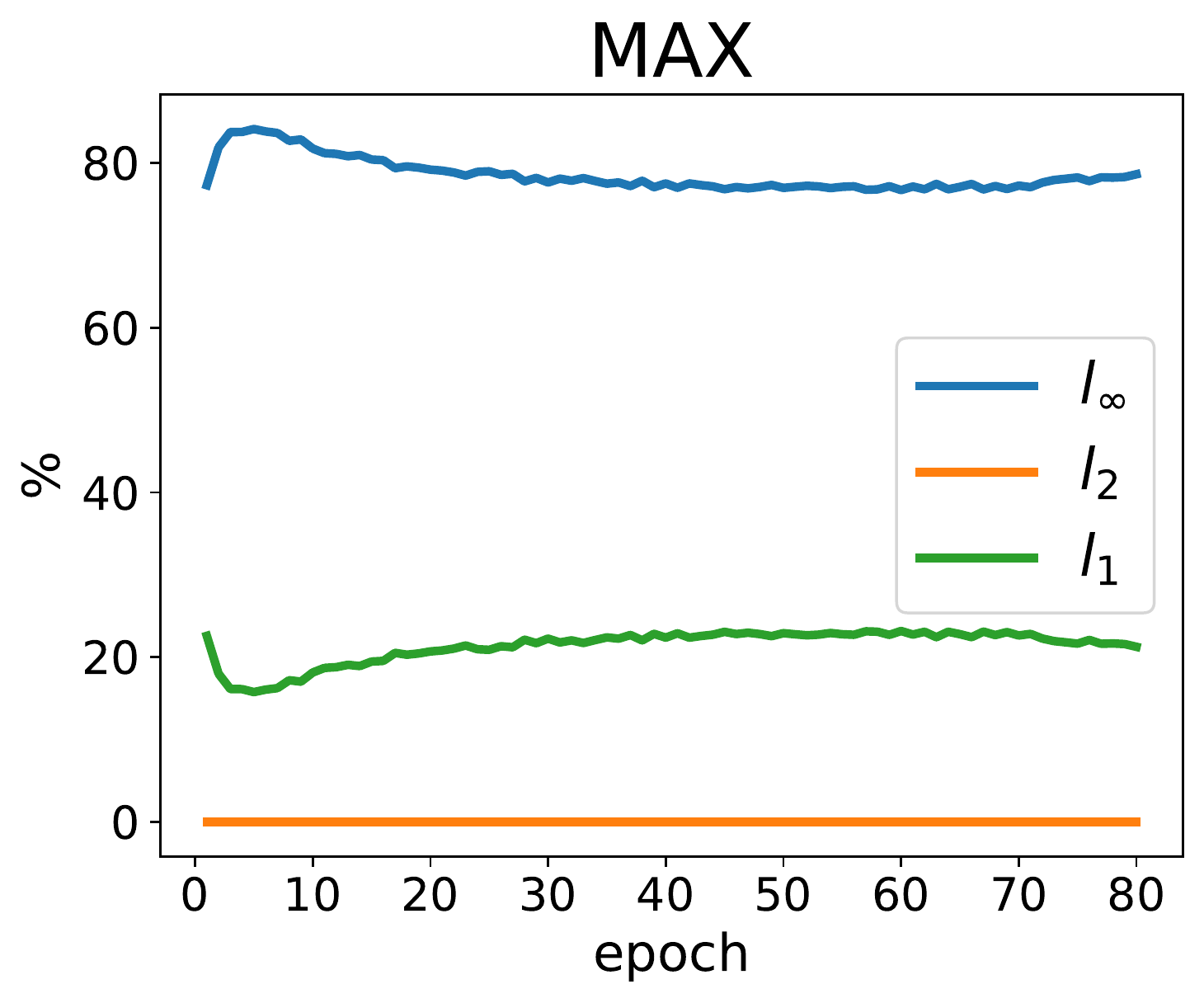} \hspace{10mm}
\includegraphics[width=.70\columnwidth]{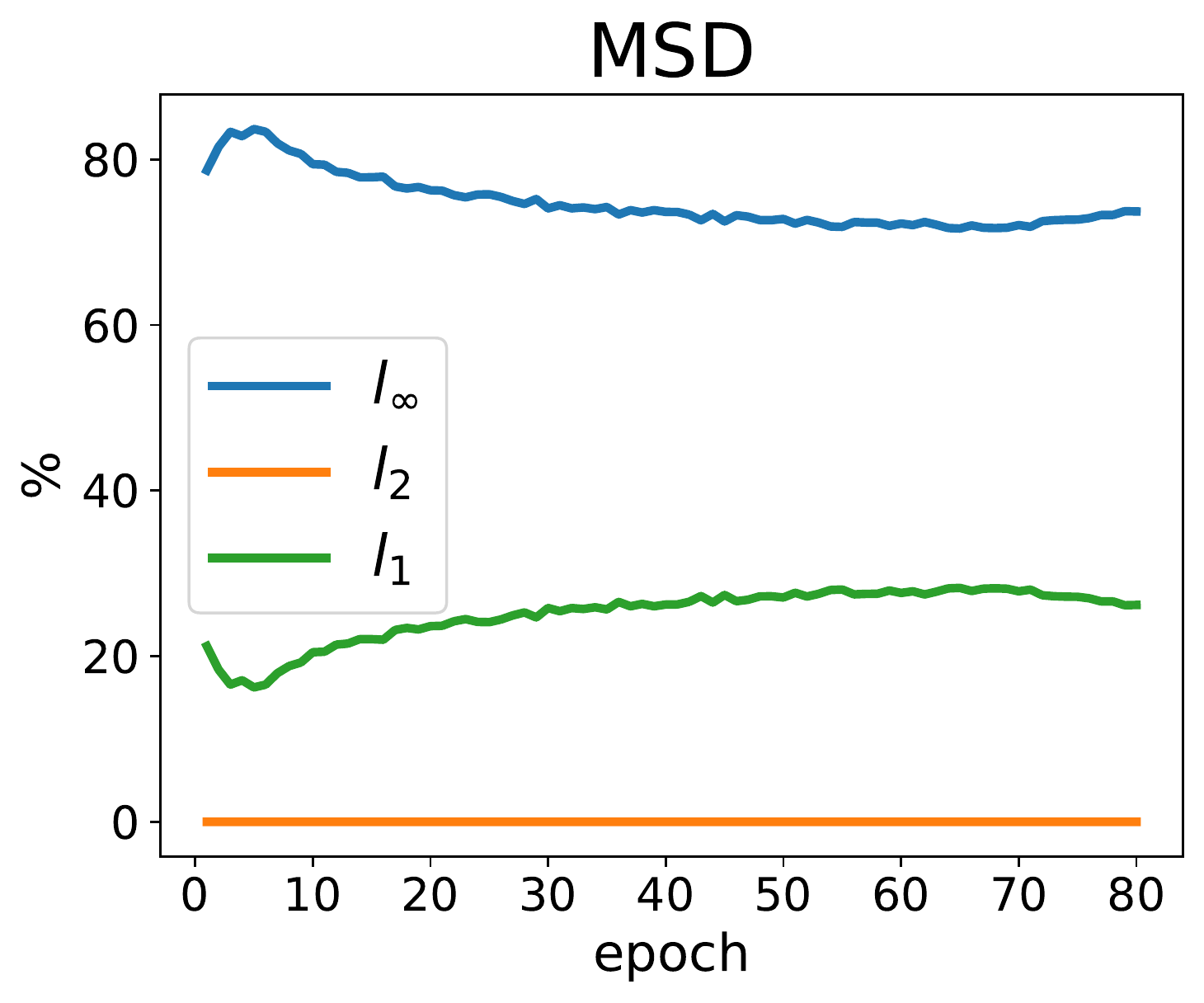}
\caption{\textbf{CIFAR-10, ResNet18.} \textbf{Left:} For MAX-training we show for each epoch during training the percentage of points attaining for the indicated $l_p$-threat model ($p \in \{1,2,\infty\}$) the highest loss over the the three threat models. \textbf{Right:} For MSD-training we show the percentage of steps taken wrt each threat model over epochs (note that MSD does the steepest descent step for each $l_p$-threat model and then realizes the one yielding maximal loss).}\label{fig:freq_threat_models}
\end{figure*}

Both MAX and MSD schemes perform adversarial training considering all the threat models simultaneously, and we  analyze here their training procedure in more detail. Fig.~\ref{fig:freq_threat_models} shows for the MAX strategy how many times, in percentage, for each epoch the points computed for each threat model realize the maximal loss over the three attacks ($l_1,l_2,l_\infty$), and then are subsequently used for the update of the model. Similarly, for MSD, we show the frequency with which a step wrt each $l_p$-norm is taken when computing the adversarial points (average over all iterations and training points). In both cases the $l_\infty$-threat model is the most used one with the $l_1$-threat model being used $3-4$ times less often. However, the $l_2$-threat model is almost never chosen. This empirically confirms the analysis from Theorem \ref{th:bounds_convex} which shows that training only wrt $l_\infty$ and $l_1$ is (at least for a linear classifier) sufficient to achieve $l_2$-robustness for the chosen $\epsilon_2$ and thus during training no extra updates wrt the $l_2$-threat model are necessary. This is in line with the results reported in Fig.~\ref{fig:curves} which show that for different thresholds the $l_2$-robustness achieved by 
E-AT-training is similar to that of standard $l_2$-training (for generating Fig.~\ref{fig:curves} we use FAB attack \citep{CroHei2019} to compute robust accuracy at varying $\epsilon_2$).

\section{Fine-Tuning with E-AT for Multiple-Norm Robustness}
\label{sec:app_exps_ft}

\subsection{Multiple-norm robustness via fast fine-tuning} \label{sec:app_ft_othermethods}
\begin{table*}[h]
\caption{\textbf{CIFAR-10 - Fine-tuning for multiple-norm robustness:} We fine-tune with different methods for multiple norms for 3 epochs the PreAct ResNet-18 robust wrt individual norms  (mean and standard deviation of the clean and robust accuracy over 5 seeds is reported). 
} \label{tab:app_ft_othermethods} \vspace{2mm}
\centering {\small 
\tabcolsep=1.5pt
\begin{tabular}{L{28mm} | *{5}{|C{20mm}} |>{\columncolor{LightCyan}}C{20mm}}
\textit{model} & \textit{clean} & $l_\infty$ ($\epsilon_\infty=\frac{8}{255}$)
&$l_2$ ($\epsilon_2=0.5$) & $l_1$ ($\epsilon_1=12$) & \textit{average} &\textit{union} \\
\toprule
RN-18 - $l_\infty$-AT & 83.7 & 48.1 & 59.8 & 7.7 & 38.5 & 7.7 \\ 
\hfill + SAT & 83.5 $\pm$ 0.2 & 43.5 $\pm$ 0.2 & 68.0 $\pm$ 0.4 & 47.4 $\pm$ 0.5 & 53.0 $\pm$ 0.2 & 41.0 $\pm$ 0.3 \\ 
\hfill + AVG & 84.2 $\pm$ 0.4 & 43.3 $\pm$ 0.4 & 68.4 $\pm$ 0.6 & 46.9 $\pm$ 0.6 & 52.9 $\pm$ 0.4 & 40.6 $\pm$ 0.4 \\ 
\hfill + MAX & 82.2 $\pm$ 0.3 & 45.2 $\pm$ 0.4 & 67.0 $\pm$ 0.7 & 46.1 $\pm$ 0.4 & 52.8 $\pm$ 0.3 & 42.2 $\pm$ 0.6 \\ 
\hfill + MSD & 82.2 $\pm$ 0.4 & 44.9 $\pm$ 0.3 & 67.1 $\pm$ 0.6 & 47.2 $\pm$ 0.6 & 53.0 $\pm$ 0.4 & 42.6 $\pm$ 0.2 \\ 
\hfill + E-AT unif. & 82.6 $\pm$ 0.6 & 44.4 $\pm$ 0.4 & 68.0 $\pm$ 0.3 & 48.5 $\pm$ 1.0 & 53.6 $\pm$ 0.4 & 42.2 $\pm$ 0.3 \\ 
\hfill + E-AT & 82.7 $\pm$ 0.4 & 44.3 $\pm$ 0.6 & 68.1 $\pm$ 0.5 & 48.7 $\pm$ 0.5 & 53.7 $\pm$ 0.3 & 42.2 $\pm$ 0.8
\\ \midrule
RN-18 - $l_2$-AT & 88.2 & 29.8 & 68.6 & 27.5 & 42.0 & 23.1 \\ 
\hfill + SAT & 86.8 $\pm$ 0.3 & 38.2 $\pm$ 0.4 & 69.6 $\pm$ 0.7 & 49.1 $\pm$ 0.6 & 52.3 $\pm$ 0.3 & 37.2 $\pm$ 0.4 \\ 
\hfill + AVG & 86.9 $\pm$ 0.2 & 37.8 $\pm$ 0.4 & 70.2 $\pm$ 0.6 & 48.3 $\pm$ 0.5 & 52.1 $\pm$ 0.4 & 36.8 $\pm$ 0.4 \\ 
\hfill + MAX & 85.1 $\pm$ 0.8 & 42.1 $\pm$ 0.3 & 69.4 $\pm$ 0.2 & 45.6 $\pm$ 0.5 & 52.4 $\pm$ 0.2 & 40.0 $\pm$ 0.3 \\ 
\hfill + MSD & 85.3 $\pm$ 0.4 & 42.0 $\pm$ 0.7 & 69.2 $\pm$ 0.2 & 44.0 $\pm$ 0.4 & 51.8 $\pm$ 0.4 & 39.0 $\pm$ 0.5 \\ 
\hfill + E-AT unif. & 85.9 $\pm$ 0.5 & 40.0 $\pm$ 0.6 & 69.4 $\pm$ 0.7 & 50.3 $\pm$ 0.4 & 53.2 $\pm$ 0.3 & 38.9 $\pm$ 0.8 \\ 
\hfill + E-AT & 85.8 $\pm$ 0.7 & 40.7 $\pm$ 0.9 & 69.5 $\pm$ 0.5 & 49.5 $\pm$ 0.5 & 53.3 $\pm$ 0.6 & 39.4 $\pm$ 0.7
\\\midrule
RN-18  - $l_1$-AT & 87.1 & 22.0 & 64.8 & 60.3 & 49.0 & 22.0 \\ 
\hfill + SAT & 85.1 $\pm$ 0.3 & 38.4 $\pm$ 0.6 & 68.3 $\pm$ 0.4 & 55.0 $\pm$ 0.7 & 53.9 $\pm$ 0.2 & 38.2 $\pm$ 0.6 \\ 
\hfill + AVG & 86.0 $\pm$ 0.2 & 38.7 $\pm$ 0.5 & 68.4 $\pm$ 0.3 & 55.6 $\pm$ 0.6 & 54.2 $\pm$ 0.2 & 38.5 $\pm$ 0.5 \\ 
\hfill + MAX & 82.2 $\pm$ 0.3 & 42.7 $\pm$ 0.5 & 66.8 $\pm$ 0.5 & 48.1 $\pm$ 0.4 & 52.5 $\pm$ 0.3 & 41.5 $\pm$ 0.4 \\ 
\hfill + MSD & 81.8 $\pm$ 0.7 & 42.7 $\pm$ 0.5 & 66.8 $\pm$ 0.4 & 47.9 $\pm$ 0.6 & 52.5 $\pm$ 0.2 & 41.5 $\pm$ 0.5 \\ 
\hfill + E-AT unif. & 83.5 $\pm$ 0.7 & 39.8 $\pm$ 0.5 & 68.0 $\pm$ 0.2 & 55.8 $\pm$ 0.5 & 54.6 $\pm$ 0.2 & 39.6 $\pm$ 0.5 \\ 
\hfill + E-AT & 83.6 $\pm$ 0.6 & 40.5 $\pm$ 0.4 & 68.1 $\pm$ 0.1 & 55.3 $\pm$ 0.4 & 54.6 $\pm$ 0.2 & 40.3 $\pm$ 0.3
\\ \bottomrule
\end{tabular} } \end{table*}

We here show that with short fine-tuning of an $l_p$-robust model it is possible to achieve competitive multiple-norm robustness for any $p \in \{\infty, 2, 1\}$. Table~\ref{tab:app_ft_othermethods} shows that all methods for robustness in the union are able to improve in 3 epochs the robustness of $l_p$-robust classifiers to multiple threat models on CIFAR-10. Moreover, one can see that the $l_\infty$-threat model is the most challenging, and fine-tuning robust models wrt it yields the best robust accuracy in the union. Moreover, E-AT outperforms SAT which has roughly the same computational cost (the training time per epoch can be found in Table~\ref{tab:training_rand_init_wrn2810_avg}). 

\subsection{Fine-tuning natural models}
\begin{table*}[h] \centering 
\caption{\textbf{CIFAR-10 - 3 epochs of 
fine-tuning with E-AT
:} We report the results of fine-tuning PreAct ResNet-18 models 
to become robust wrt the union of the threat models. 
Fine-tuning any $l_p$-robust model leads to competitive clean and robust accuracy to full training, differently from using a naturally trained model. }\label{tab:app_ft_small_models} \vspace{2mm}
{\small \centering
\tabcolsep=3pt
\begin{tabular}{L{30mm} | *{4}{| *{2}{C{8mm}}}
|*{2}{>{\columncolor{LightCyan}}C{8mm}}}
\textit{model} & \textit{clean} && \multicolumn{2}{l}{$l_\infty$ ($\epsilon_\infty=\frac{8}{255}$)} &\multicolumn{2}{|l}{$l_2$ ($\epsilon_2=0.5$)} & \multicolumn{2}{|l|}{$l_1$ ($\epsilon_1=12$)} & \textit{union} & \\
\toprule
RN-18 - standard & 94.4 & & 0.0 & & 0.0 & & 0.0 & & 0.0 &\\ 
  \hfill + FT & 66.6 & \textcolor{red}{-27.8} & 29.9 & \textcolor{blue}{29.9} & 50.1 & \textcolor{blue}{50.1} & 38.5 & \textcolor{blue}{38.5}&29.8 & \textcolor{blue}{29.8} \\ \midrule
RN-18 - $l_\infty$ & 83.7 & & 48.1 & & 59.8 & & 7.7 & & 7.7 &\\
 \hfill + FT & 82.3 & \textcolor{red}{-1.4} & 43.4 & \textcolor{red}{-4.7} & 68.0 & \textcolor{blue}{8.2} & 48.0 & \textcolor{blue}{40.3} & 41.2 & \textcolor{blue}{33.5}\\
 \midrule
RN-18 - $l_2$ & 88.2 & & 29.8 & & 68.6 & & 27.5 & & 23.1 &\\
 \hfill + FT & 85.4 & \textcolor{red}{-2.8} & 40.6 & \textcolor{blue}{10.8} & 69.8 & \textcolor{blue}{1.2} & 48.7 & \textcolor{blue}{21.2} & 39.1 & \textcolor{blue}{16.0}\\
 \midrule
RN-18 
- $l_1$ & 87.1 & & 22.0 & & 64.8 & & 60.3 & & 22.0 &\\
 
 \hfill + FT & 83.5 & \textcolor{red}{-3.6} & 40.3 & \textcolor{blue}{18.3} & 68.1 & \textcolor{blue}{3.3} & 55.7 & \textcolor{red}{-4.6} & 40.1 & \textcolor{blue}{18.1}
\\
\bottomrule
\end{tabular} }
\end{table*}

We fine-tune with E-AT for 3 epochs PreAct ResNet-18 either naturally trained or robust wrt a single $l_p$-norm. Table~\ref{tab:app_ft_small_models} shows clean and robust accuracy for each threat model for the initial classifier and after E-AT fine-tuning: while for all robust models the fine-tuning yields values competitive with the full training for multiple norms (see Table~\ref{tab:training_rand_init_wrn2810_avg}), starting from a standard model leads to significantly lower both clean performance and robustness in the union of the three $l_p$-balls.

\subsection{Runtime with large models}
We reported in Table~\ref{tab:ft_othermethods} and Table~\ref{tab:training_rand_init_wrn2810_avg} the runtime per epoch of E-AT. For larger architectures the computational cost increases significantly, and adversarial training with the WideResNet-70-16, the largest one we consider, on CIFAR-10 takes, in our experiments, around 6100 s per epoch when using only the training set and over 10000 s if the unlabelled data is used (since twice more training steps are effectively used). Moreover, fine-tuning on ImageNet takes around 6 h per epoch for $l_\infty$ and $l_2$ (5 steps of adversarial training), 20 h per epoch for $l_1$ (15 steps), 13.5 h per epoch with E-AT (with the architectures we consider). This shows how transferring robustness with fine-tuning might allow to obtain classifiers robust wrt different threat models fast and at much lower computational cost.

\subsection{
Effect of biased sampling scheme}\label{sec:app_abl_seed_sampling}
We compare the biased sampling scheme of E-AT introduced in Eq.~\eqref{eq:probl1linf} to a uniform one, where one chooses uniformly at random which $l_p$, with $p\in\{1, \infty\}$, attack to use for adversarial training for each batch. 
This is referred to as E-AT unif. in Table~\ref{tab:app_ft_othermethods}, where one case see that the biased sampling schemes yields slightly better results when fine-tuning the $l_2$- and in particular the $l_1$-robust model: we hypothesize that, since $l_\infty$ is the most challenging threat model, it is important to use it more often at training time when the initial model is non robust wrt $l_\infty$. Moreover, we test E-AT unif. for full training (see Table~\ref{tab:training_rand_init_wrn2810_avg}).

\subsection{Results using different number of epochs}
\begin{table*}[h]
\caption{\textbf{CIFAR-10 - Fine-tuning for more epochs:} We show the effect of 
fine-tuning for different number of epochs (3 is the standard we use) the PreAct ResNet-18 (standard or robust wrt $l_\infty$). 
} \label{tab:app_ft_epochs} \vspace{2mm}
\centering {\small
\tabcolsep=2.5pt
\begin{tabular}{L{35mm} | *{4}{| *{2}{C{8mm}}} | *{2}{>{\columncolor{LightCyan}}C{8mm}}}
\textit{model} & \textit{clean} && \multicolumn{2}{l}{$l_\infty$ ($\epsilon_\infty=\frac{8}{255}$)} &\multicolumn{2}{|l}{$l_2$ ($\epsilon_2=0.5$)} & \multicolumn{2}{|l|}{$l_1$ ($\epsilon_1=12$)} & \textit{union} & \\
\toprule
RN-18 - $l_\infty$ &83.7 & & 48.1 & & 59.8 & & 7.7 & & 7.7 &\\
 \hfill + 3 epochs FT & 82.3 & \textcolor{red}{-1.4} & 43.4 & \textcolor{red}{-4.7} & 68.0 & \textcolor{blue}{8.2} & 48.0 & \textcolor{blue}{40.3} & 41.2 & \textcolor{blue}{33.5}\\
 \hfill + 5 epochs FT & 83.0 & \textcolor{red}{-0.7} & 45.2 & \textcolor{red}{-2.9} & 68.8 & \textcolor{blue}{9.0} & 50.1 & \textcolor{blue}{42.4} & 43.1 & \textcolor{blue}{35.4}\\
 \hfill + 7 epochs FT & 83.1 & \textcolor{red}{-0.6} & 44.6 & \textcolor{red}{-3.5} & 68.7 & \textcolor{blue}{8.9} & 50.4 & \textcolor{blue}{42.7} & 42.6 & \textcolor{blue}{34.9}\\
 \hfill + 10 epochs FT & 84.0 & \textcolor{blue}{0.3} & 44.9 & \textcolor{red}{-3.2} & 69.2 & \textcolor{blue}{9.4} & 51.0 & \textcolor{blue}{43.3} & 42.8 & \textcolor{blue}{35.1}\\
 \hfill + 15 epochs FT & 84.6 & \textcolor{blue}{0.9} & 44.9 & \textcolor{red}{-3.2} & 69.5 & \textcolor{blue}{9.7} & 52.1 & \textcolor{blue}{44.4} & 43.2 & \textcolor{blue}{35.5}\\ \midrule
RN-18 - standard & 94.4 & & 0.0 & & 0.0 & & 0.0 & & 0.0 &\\ 
 \hfill + 3 epochs FT & 66.6 & \textcolor{red}{-27.8} & 29.9 & \textcolor{blue}{29.9} & 50.1 & \textcolor{blue}{50.1} & 38.5 & \textcolor{blue}{38.5} & 29.8 & \textcolor{blue}{29.8}\\ 
 \hfill + 5 epochs FT & 70.6 & \textcolor{red}{-23.8} & 33.8 & \textcolor{blue}{33.8} & 55.2 & \textcolor{blue}{55.2} & 44.4 & \textcolor{blue}{44.4} & 33.4 & \textcolor{blue}{33.4}\\ 
 \hfill + 7 epochs FT & 72.1 & \textcolor{red}{-22.3} & 36.1 & \textcolor{blue}{36.1} & 58.9 & \textcolor{blue}{58.9} & 45.9 & \textcolor{blue}{45.9} & 35.6 & \textcolor{blue}{35.6}\\ 
 \hfill + 10 epochs FT & 75.4 & \textcolor{red}{-19.0} & 37.1 & \textcolor{blue}{37.1} & 61.0 & \textcolor{blue}{61.0} & 47.9 & \textcolor{blue}{47.9} & 36.9 & \textcolor{blue}{36.9}\\ 
 \hfill + 15 epochs FT & 76.0 & \textcolor{red}{-18.4} & 40.2 & \textcolor{blue}{40.2} & 61.6 & \textcolor{blue}{61.6} & 49.2 & \textcolor{blue}{49.2} & 40.0 & \textcolor{blue}{40.0}
\\ \bottomrule
\end{tabular} }
\end{table*}
Table~\ref{tab:app_ft_epochs} shows the effect of our 
E-AT-fine-tuning for different numbers of epochs on a RN-18 either robust wrt $l_\infty$ or naturally trained. For the robust model, with longer training the clean accuracy progressively improves, as well as the robustness in the union of the threat models. In particular, the models fine-tuned for 15 epochs has robust accuracy similar to that achieved by the MAX-training ($43.2\%$ compared to $43.3\%$, see Table~\ref{tab:training_rand_init_wrn2810_avg}), while still being significantly faster even considering the training time of the initial model. When starting from a standard model, E-AT fine-tuning leads to a large drop in clean accuracy while the robustness in the union remains lower than what can be achieved with robust models, even when using 15 epochs. Similarly, we fine-tune for 3 epochs, instead of 1 as done above, the $l_2$-robust model on ImageNet from \citet{robustness} with our E-AT. Table~\ref{tab:ImageNet-fine-moreepochs} shows that the longer fine-tuning improves all the performance metrics between 0.6\% and 1.3\%. Moreover, since we use a single epoch of fine-tuning on ImageNet, we test its effect on CIFAR-10. In Table~\ref{tab:app_ft_singleepoch} we fine-tune with E-AT models adversarially trained wrt a single norm for 1 epoch: this is sufficient to significantly increase the robustness in the union of the threat models, which gets close to that obtained with the standard 3 epochs (differences are in the range 0.6\% to 2.5\%). In particular, the $l_\infty$-robust classifier is again the most suitable for the fine-tuning, since it has been trained in the most challenging threat model.

\begin{table*}[h] \centering \caption{\label{tab:ImageNet-fine-moreepochs}
\textbf{ImageNet - Fine-tuning for more epochs:} 
We fine-tune the $l_2$-robust model from \cite{robustness} for either 1 or 3 epochs with our E-AT scheme.}
\vspace{2mm}
{\small
\tabcolsep=2.5pt
\begin{tabular}{L{50mm} | *{4}{| *{2}{C{8mm}}} |*{2}{>{\columncolor{LightCyan}}C{8mm}}}
\textit{model} & \textit{clean} && \multicolumn{2}{|l}{$l_\infty$ ($\epsilon_\infty=\frac{4}{255}$)} &\multicolumn{2}{|l}{$l_2$ ($\epsilon_2=2$)} & \multicolumn{2}{|l|}{$l_1$ ($\epsilon_1=255$)} & 
union & \\
\toprule
RN-50 - $l_2$  \citep{robustness}& 58.7 & & 25.0 & & 40.5 & & 14.0 & & 13.5 &\\ 
 \hfill + 1 epochs FT & 56.7 & \textcolor{red}{-2.0} & 26.7 & \textcolor{blue}{1.7} & 41.0 & \textcolor{blue}{0.5} & 25.4 & \textcolor{blue}{11.4} & 23.1 & \textcolor{blue}{9.6}\\ 
 \hfill + 3 epochs FT & 57.4 & \textcolor{red}{-1.3} & 27.8 & \textcolor{blue}{2.8} & 41.6 & \textcolor{blue}{1.1} & 26.7 & \textcolor{blue}{12.7} & 23.7 & \textcolor{blue}{10.2}
\\ \bottomrule
\end{tabular} } \end{table*}

\begin{table*}[h]
\caption{\textbf{CIFAR-10 - Fine-tuning for 1 epoch:} We show the effect of 
fine-tuning for a single (compared to the standard 3) models robust wrt a single norm.
} \label{tab:app_ft_singleepoch} \vspace{2mm}
\centering {\small
\tabcolsep=2.5pt
\begin{tabular}{L{50mm} | *{4}{| *{2}{C{8mm}}} | *{2}{>{\columncolor{LightCyan}}C{8mm}}}
\textit{model} & \textit{clean} && \multicolumn{2}{l}{$l_\infty$ ($\epsilon_\infty=\frac{8}{255}$)} &\multicolumn{2}{|l}{$l_2$ ($\epsilon_2=0.5$)} & \multicolumn{2}{|l|}{$l_1$ ($\epsilon_1=12$)} & \textit{union} & \\
\toprule
RN-50 \citep{robustness} - $l_\infty$ & 88.7 & & 50.9 & & 59.4 & & 5.0 & & 5.0 &\\ 
 \hfill + 1 epochs FT & 84.8 & \textcolor{red}{-3.9} & 46.6 & \textcolor{red}{-4.3} & 68.3 & \textcolor{blue}{8.9} & 47.2 & \textcolor{blue}{42.2} & 42.8 & \textcolor{blue}{37.8}\\ 
 \hfill + 3 epochs FT & 86.2 & \textcolor{red}{-2.5} & 46.0 & \textcolor{red}{-4.9} & 70.1 & \textcolor{blue}{10.7} & 49.2 & \textcolor{blue}{44.2} & 43.4 & \textcolor{blue}{38.4}\\ \midrule 
RN-50 \citep{robustness} - $l_2$ & 91.5 & & 29.7 & & 70.3 & & 27.0 & & 23.0 &\\ 
 \hfill + 1 epochs FT & 85.9 & \textcolor{red}{-5.6} & 41.8 & \textcolor{blue}{12.1} & 69.6 & \textcolor{red}{-0.7} & 47.6 & \textcolor{blue}{20.6} & 39.7 & \textcolor{blue}{16.7}\\ 
 \hfill + 3 epochs FT & 87.8 & \textcolor{red}{-3.7} & 43.1 & \textcolor{blue}{13.4} & 70.8 & \textcolor{blue}{0.5} & 50.2 & \textcolor{blue}{23.2} & 41.7 & \textcolor{blue}{18.7}\\ \midrule 
RN-18 \citep{croce2021mind} - $l_1$ & 87.1 & & 22.0 & & 64.8 & & 60.3 & & 22.0 &\\ 
 \hfill + 1 epochs FT & 78.9 & \textcolor{red}{-8.2} & 37.7 & \textcolor{blue}{15.7} & 62.9 & \textcolor{red}{-1.9} & 51.3 & \textcolor{red}{-9.0} & 37.6 & \textcolor{blue}{15.6}\\ 
 \hfill + 3 epochs FT & 83.5 & \textcolor{red}{-3.6} & 40.3 & \textcolor{blue}{18.3} & 68.1 & \textcolor{blue}{3.3} & 55.7 & \textcolor{red}{-4.6} & 40.1 & \textcolor{blue}{18.1}
\\ \bottomrule
\end{tabular} } \end{table*}

\subsection{Fine-tuning perceptually robust models}
\begin{table*}[h] \centering \caption{\label{tab:CIFAR10-rn50}\textbf{CIFAR-10 - E-AT fine-tuning of perceptually robust models:} 
We use E-AT to fine-tune for 3 epochs the PAT model, robust wrt LPIPS, and compare to fine-tuning $l_p$-robust models. 
} \vspace{2mm}
{\small
\tabcolsep=2.5pt
\begin{tabular}{L{37mm} | *{4}{| *{2}{C{8mm}}} 
|*{2}{>{\columncolor{LightCyan}}C{8mm}}}
\textit{model} & \textit{clean} && \multicolumn{2}{|l}{$l_\infty$ ($\epsilon_\infty=\frac{8}{255}$)} &\multicolumn{2}{|l}{$l_2$ ($\epsilon_2=0.5$)} & \multicolumn{2}{|l|}{$l_1$ ($\epsilon_1=12$)} & 
\textit{union} & \\
\toprule
RN-50  - $l_\infty$ & 88.7 & & 50.9 & & 59.4 & & 5.0 & & 5.0 &\\ 
 \citep{robustness} \hfill + FT & 86.2 & \textcolor{red}{-2.5} & 46.0 & \textcolor{red}{-4.9} & 70.1 & \textcolor{blue}{10.7} & 49.2 & \textcolor{blue}{44.2} & 43.4 & \textcolor{blue}{38.4}\\ 
 \midrule
RN-50  - $l_2$ & 91.5 & & 29.7 & & 70.3 & & 27.0 & & 23.0 &\\ 
 \citep{robustness} \hfill + FT & 87.8 & \textcolor{red}{-3.7} & 43.1 & \textcolor{blue}{13.4} & 70.8 & \textcolor{blue}{0.5} & 50.2 & \textcolor{blue}{23.2} & 41.7 & \textcolor{blue}{18.7}\\ 
 \midrule
RN-50  - PAT 
& 82.6 & & 31.1 & & 62.4 & & 33.6 & & 27.7 &\\ 
 \citep{laidlaw2021perceptual} \hfill + FT & 83.7 & \textcolor{blue}{1.1} & 43.7 & \textcolor{blue}{12.6} & 68.5 & \textcolor{blue}{6.1} & 50.7 & \textcolor{blue}{17.1} & 42.3 & \textcolor{blue}{14.6}
\\ \bottomrule
\end{tabular} } \end{table*}

We test the effect of E-AT fine-tuning on a model trained to be robust to perturbations which are aligned with human perception. In particular, we use the classifier obtained with perceptual adversarial training (PAT), i.e. wrt the LPIPS metric, from \citet{laidlaw2021perceptual}, and compare it to two models with the same architecture (ResNet-50) adversarially trained wrt $l_\infty$ and $l_2$. Table~\ref{tab:CIFAR10-rn50} shows the robustness in every threat model for the original models and those obtained with 3 epochs of E-AT fine-tuning. The PAT classifier has initially the highest robustness in the union, confirming the obersevation of \citet{laidlaw2021perceptual} that PAT provides some robustness to unseen attacks. After fine-tuning, all three models achieve similar worst-case robustness, with the classifier originally $l_\infty$-robust being slightly better. This shows that our E-AT fine-tuning is effective even when applied to models adversarially trained not wrt an $l_p$-norm.

\subsection{Full training for multiple norm robustness} \label{sec:app_fulltraining}
\begin{table*}[h] \centering \caption{\textbf{CIFAR-10 - Comparison of different full training schemes:} We compare methods for multiple-norm robustness on training from random initialization.
\label{tab:training_rand_init_wrn2810_avg}}
\vspace{+2mm}
{\small
\tabcolsep=2pt
\begin{tabular}{L{18mm} || C{18mm} *{4}{|C{20mm}} 
*{1}{|>{\columncolor{LightCyan}}C{20mm}}
|>{\columncolor{LightCyan}}C{15mm}}
\textit{method} & \textit{clean} & $l_\infty$ ($\epsilon_\infty=\frac{8}{255}$) &$l_2$ ($\epsilon_2=0.5$) & $l_1$ ($\epsilon_1=12$) &\textit{average}& \textit{union} & \textit{time/epoch} \\ \toprule
\multicolumn{7}{c}{}\\ \multicolumn{7}{l}{\textbf{WideResNet-28-10}}\\ \toprule
$l_\infty$-AT & 82.6 $\pm$ 0.5 & 52.0 $\pm$ 0.7 & 59.7 $\pm$ 0.2 & 9.1 $\pm$ 0.2 & 40.3 $\pm$ 0.4 & 9.1 $\pm$ 0.2 & 922 s \\ 
$l_2$-AT & 88.2 $\pm$ 0.4 & 35.9 $\pm$ 0.2 & 70.9 $\pm$ 0.4 & 36.1 $\pm$ 0.2 & 47.6 $\pm$ 0.2 & 31.3 $\pm$ 0.2 & 928 s \\ 
$l_1$-AT & 83.7 $\pm$ 0.2 & 30.7 $\pm$ 0.7 & 65.1 $\pm$ 0.5 & 61.6 $\pm$ 0.3 & 52.5 $\pm$ 0.5 & 30.7 $\pm$ 0.7 & 949 s \\ \midrule
SAT & 80.5 $\pm$ 0.6 & 45.9 $\pm$ 0.5 & 66.7 $\pm$ 0.3 & 55.9 $\pm$ 0.5 & 56.2 $\pm$ 0.4 & 45.7 $\pm$ 0.6 & 925 s \\ 
MNG-AC & 81.3 $\pm$ 0.3 & 43.5 $\pm$ 0.7 & 66.9 $\pm$ 0.2 & 57.6 $\pm$ 0.8 & 56.0 $\pm$ 0.4 & 43.3 $\pm$ 0.7 & 1500 s \\ 
AVG & 82.5 $\pm$ 0.4 & 45.4 $\pm$ 1.1 & 68.0 $\pm$ 0.9 & 55.0 $\pm$ 0.2 & 56.1 $\pm$ 0.7 & 45.1 $\pm$ 1.1 & 2771 s \\ 
MAX & 79.9 $\pm$ 0.1 & 48.4 $\pm$ 0.7 & 65.3 $\pm$ 0.3 & 50.2 $\pm$ 0.6 & 54.6 $\pm$ 0.5 & 47.4 $\pm$ 0.8 & 2479 s \\ 
MSD & 80.6 $\pm$ 0.3 & 48.0 $\pm$ 0.2 & 65.6 $\pm$ 0.3 & 51.7 $\pm$ 0.4 & 55.1 $\pm$ 0.2 & 46.9 $\pm$ 0.1 & 1554 s \\ 
E-AT unif. & 79.7 $\pm$ 0.2 & 45.4 $\pm$ 0.5 & 66.0 $\pm$ 0.5 & 55.6 $\pm$ 0.5 & 55.7 $\pm$ 0.4 & 45.1 $\pm$ 0.7 & 939 s \\ 
E-AT & 79.9 $\pm$ 0.7 & 46.6 $\pm$ 0.2 & 66.2 $\pm$ 0.6 & 56.0 $\pm$ 0.4 & 56.3 $\pm$ 0.3 & 46.4 $\pm$ 0.3 & 921 s
\\ \midrule
\multicolumn{7}{c}{}\\ \multicolumn{7}{l}{\textbf{PreAct ResNet-18}} \\ \toprule
$l_\infty$-AT & 84.0 $\pm$ 0.3 & 48.1 $\pm$ 0.2 & 59.7 $\pm$ 0.4 & 6.3 $\pm$ 1.0 & 38.0 $\pm$ 0.3 & 6.3 $\pm$ 1.0 & 151 s \\ 
$l_2$-AT & 88.9 $\pm$ 0.6 & 27.3 $\pm$ 1.8 & 68.7 $\pm$ 0.1 & 25.3 $\pm$ 1.6 & 40.5 $\pm$ 1.1 & 20.9 $\pm$ 1.8 & 153 s \\ 
$l_1$-AT & 85.9 $\pm$ 1.1 & 22.1 $\pm$ 0.1 & 64.9 $\pm$ 0.5 & 59.5 $\pm$ 0.8 & 48.8 $\pm$ 0.4 & 22.1 $\pm$ 0.1 & 195 s \\  \midrule
SAT & 83.9 $\pm$ 0.8 & 40.7 $\pm$ 0.7 & 68.0 $\pm$ 0.4 & 54.0 $\pm$ 1.2 & 54.2 $\pm$ 0.8 & 40.4 $\pm$ 0.7 & 161 s \\ 
AVG & 84.6 $\pm$ 0.3 & 40.8 $\pm$ 0.7 & 68.4 $\pm$ 0.7 & 52.1 $\pm$ 0.4 & 53.8 $\pm$ 0.1 & 40.1 $\pm$ 0.8 & 479 s \\ 
MAX & 80.4 $\pm$ 0.5 & 45.7 $\pm$ 0.9 & 66.0 $\pm$ 0.4 & 48.6 $\pm$ 0.8 & 53.4 $\pm$ 0.5 & 44.0 $\pm$ 0.7 & 466 s \\ 
MSD (*) & 82.1 & 43.1 & 64.5 & 46.5 & 51.4 & 41.4 & - \\ 
MSD & 81.1 $\pm$ 1.1 & 44.9 $\pm$ 0.6 & 65.9 $\pm$ 0.6 & 49.5 $\pm$ 1.2 & 53.4 $\pm$ 0.4 & 43.9 $\pm$ 0.8 & 306 s \\ 
E-AT unif. & 82.2 $\pm$ 1.8 & 42.7 $\pm$ 0.7 & 67.5 $\pm$ 0.5 & 53.6 $\pm$ 0.1 & 54.6 $\pm$ 0.2 & 42.4 $\pm$ 0.6 & 163 s \\ 
E-AT & 81.9 $\pm$ 1.4 & 43.0 $\pm$ 0.9 & 66.4 $\pm$ 0.6 & 53.0 $\pm$ 0.3 & 54.2 $\pm$ 0.4 & 42.4 $\pm$ 0.7 & 160 s
\\ \bottomrule
\end{tabular} } \end{table*}
We report in Table~\ref{tab:training_rand_init_wrn2810_avg} the detailed results
of training for multiple norms robustness from random initialization with PreAct ResNet-18 \citep{he2016identity} and WideResNet-28-10 \citep{ZagKom2016}. 
For both architectures MAX and MSD attain the best results in the union. However, E-AT is only slightly worse while being 2-3 times less expensive. 
We include MNG-AC from \citet{madaan2020learning} for the larger architecture as that is used also in the original paper. Moreover, since it has PreAct ResNet-18 as architecture, we additionally include the original MSD model of \citet{maini2020adversarial}, marked with (*), which obtains $41.4\%$ robustness in the union, whereas with our reimplementation we get $43.9\%$, improving their results significantly. Note that they reported in their paper $47.0\%$ robustness in the union, while our APGD-based evaluation reduces this to 41.4\% which shows that our robustness evaluation is significantly stronger. Finally, we add the results of E-AT without biased sampling scheme (E-AT unif.): 
this achieves worse results than E-AT on the WRN-28-10, showing the effectiveness of the biased sampling.

\subsection{Robustness against unseen non $l_p$-bounded attacks} \label{sec:app_otherattacks}
To test robustness against $l_0$-attacks we use Sparse-RS \citep{croce2020sparsers} with a budget of 18 pixels and 10k queries. We adopt patches of size $5\times 5$ pixels, optimized with the PGD-based attack from \citet{rao2020adversarial} (without constraints on the position of the patch on the images), and frames of width 1 pixel \citep{zajac2019adversarial}, again optimized with PGD: in both cases we use 10 random restarts of 100 iterations. For the adversarial corruptions we use the original implementation \citep{kang2019testing} with 100 iterations and search for a budget $\epsilon$ for which the models show different levels of robustness (in details, for fog $\epsilon=128$, snow $\epsilon=0.5$, Gabor noise $\epsilon=60$, elastic $\epsilon=0.125$, $l_\infty$-JPEG $\epsilon=0.25$). Finally, we average the classification accuracy over the 5 severities of the common corruptions. All the statistics are on 1000 test points.

\section{Fine-tuning to a different threat model: additional results}\label{sec:app_indiv_threat_models}
\begin{table}[H]\caption{\label{tab:imagenet-fine-lp}\textbf{Fine-tuning $l_p$-robust models to another threat model:} we fine-tune for 1 epoch models trained wrt $l_p$ with adversarial training wrt another $l_q$.} \vspace{2mm}
\centering {\small
\begin{tabular}[t]{L{20mm} || C{10mm}
*{3}{|*{1}{C{10mm}}} }
\multicolumn{5}{c}{\textbf{ImageNet}} \\
& \multirow{1}{*}{\textit{clean}} & $l_\infty$ & $l_2$ & $l_1$\\
\toprule
\multicolumn{5}{l}{RN-50 \citep{robustness} - $l_\infty$} \\ \toprule
original & 62.9  & \cellcolor{blue!10}29.8  & 17.7  & 0.0 \\
 \hfill + FT wrt $l_2$ & 62.9 & 25.5 & \cellcolor{orange!10} 41.5 & 8.4\\
 \hfill + FT wrt $l_1$ & 57.7 & 18.0 & 37.6 & \cellcolor{orange!10}27.4
\\\bottomrule
\end{tabular}}\end{table}
We additionally report in Table~\ref{tab:imagenet-fine-lp} the results of fine-tuning for 1 epoch the RN-50 \citep{robustness} robust wrt $l_\infty$ with adversarial training wrt others $l_q$-threat models: similarly to the other models this is sufficient to obtain significant robustness in the new threat model.

\section{Experiments on MNIST}\label{sec:app_mnist}
\begin{table*}[h]
\caption{\textbf{MNIST - Comparison of full training schemes and fine-tuning with E-AT for multiple norm robustness:} We train classifier (architecture as in \citet{maini2020adversarial}) on MNIST with different training scheme. For SAT and E-AT we report, together with the statistics over multiple random seeds, the results of the best run. Additionally, we show the results of fine-tuning the $l_2$-AT model with E-AT for different numbers of epochs, which achieves the best results.
(*) AVG, MAX and MSD classifiers are those provided by \citet{maini2020adversarial}.
} \label{tab:mnist}\vspace{2mm}
\centering {\small 
\tabcolsep=2pt
\begin{tabular}{L{30mm} || C{20mm} *{3}{|C{20mm}} |>{\columncolor{LightCyan}}C{20mm}}
\textit{model} & \textit{clean} & $l_\infty$ ($\epsilon_\infty=0.3$)
&$l_2$ ($\epsilon_2=2$) & $l_1$ ($\epsilon_1=10$) & \textit{union} \\
\toprule
$l_\infty$-AT & 98.9 $\pm$ 0.12 & 90.0 $\pm$ 0.45 & 8.4 $\pm$ 1.49 & 6.0 $\pm$ 0.65 & 4.2 $\pm$ 0.73 \\ 
$l_2$-AT & 98.8 $\pm$ 0.17 & 0.0 $\pm$ 0.05 & 70.7 $\pm$ 0.26 & 59.1 $\pm$ 0.37 & 0.0 $\pm$ 0.05 \\ 
$l_1$-AT & 98.8 $\pm$ 0.09 & 0.0 $\pm$ 0.00 & 45.8 $\pm$ 0.42 & 77.2 $\pm$ 0.14 & 0.0 $\pm$ 0.00 \\ \midrule 
SAT & 98.6 $\pm$ 0.17 & 62.0 $\pm$ 0.86 & 65.7 $\pm$ 1.69 & 61.3 $\pm$ 1.29 & 53.9 $\pm$ 1.25 \\ 
AVG (*) & 99.1 & 58.6 & 60.8 & 22.5 & 21.1 \\ 
MAX (*) & 98.6 & 39.4 & 59.9 & 25.6 & 20.3 \\ 
MSD (*) & 98.2 & 63.7 & 66.6 & 51.0 & 48.7 \\ 
E-AT unif. & 98.8 $\pm$ 0.12 & 67.1 $\pm$ 3.03 & 50.2 $\pm$ 4.93 & 62.0 $\pm$ 4.59 & 45.9 $\pm$ 4.43 \\
E-AT & 98.7 $\pm$ 0.12 & 69.0 $\pm$ 3.66 & 56.4 $\pm$ 3.94 & 61.1 $\pm$ 4.03 & 50.6 $\pm$ 3.89 \\
$l_2$-AT + E-AT (3 ep.) & 96.9 $\pm$ 0.32 & 57.5 $\pm$ 0.92 & 67.8 $\pm$ 0.58 & 62.0 $\pm$ 1.16 & 54.6 $\pm$ 0.59 \\ 
$l_2$-AT + E-AT (5 ep.) & 97.4 $\pm$ 0.21 & 60.6 $\pm$ 2.19 & 65.9 $\pm$ 0.56 & 63.8 $\pm$ 0.93 & 56.2 $\pm$ 1.16 \\ \midrule
\multicolumn{6}{c}{} \\ \multicolumn{6}{l}{\textbf{best run}} \\ \toprule
SAT  & 98.8 & 62.7 & 67.2 & 62.6 & 55.3 \\
E-AT unif.   & 98.9 & 67.4 & 56.3 & 63.6 & 51.6 \\
E-AT   & 98.8 & 71.0 & 58.4 & 62.0 & 54.4 \\ 
$l_2$-AT + E-AT (3 ep.)   & 97.3 & 58.0 & 67.9 & 62.9 & 55.7 \\ 
$l_2$-AT + E-AT (5 ep.)   & 97.5 & 61.6 & 66.2 & 64.0 & 57.5
\\ \bottomrule
\end{tabular} } \end{table*}
We further test the different techniques on the MNIST dataset. We use the same CNN of \citet{maini2020adversarial} as architecture and $\epsilon_\infty=0.3$, $\epsilon_2=2$ and $\epsilon_1=10$ as thresholds at which evaluating robustness, as done by \citet{maini2020adversarial}. We note that while it is an easier dataset, MNIST is challenging when it comes to adversarial training since it presents unexpected phenomena: e.g. \citet{TraBon2019} noted that $l_\infty$-adversarial training induces gradient obfuscation when using attack wrt $l_2$ and $l_1$, and both \citet{TraBon2019, maini2020adversarial} had to use many PGD-steps (up to 100), and \citet{TraBon2019} even a ramp-up schedule for the $\epsilon$ during training. While other modifications to the training setup might be beneficial for some or all the methods, we 
just increased the number of APGD-steps to 50 for $l_1$ (see more details below). In Table~\ref{tab:mnist} we compare E-AT to SAT, AVG, MAX and MSD (for the last three we use the models provided by \citet{maini2020adversarial}). First, E-AT outperforms the available classifiers trained with AVG, MAX and MSD, meaning that even on MNIST it is a strong baseline. However, in this case, SAT, which trains on all types of perturbations, achieves better results than E-AT on average: E-AT has higher variance over runs but the best run (over multiple seeds) is close to the best one of SAT in terms of robustness in the union (55.3\% vs 54.4\%). Interestingly, SAT has much higher robustness wrt $l_2$ compared to E-AT, but this is somehow expected since Eq.~\eqref{eq:bounds_convex} would ``predict’’ robustness for E-AT at $\epsilon_2 \approx 1.7$ while $\epsilon_2=2$ is used for testing, and this is precise only for linear models. Thus the slightly worse performance of E-AT compared to SAT for the chosen radii of the threat models is to be expected from our geometric analysis.

Moreover, since we have shown that fine-tuning an $l_p$-robust model with E-AT yields high multiple norms robustness, and given that E-AT from random initialization is weak mostly wrt $l_2$, we fine-tune the $l_2$-AT classifier with E-AT. 
This, with just 3 or 5 epochs, significantly outperforms SAT (up to +2.3\% robustness in the union), while preserving $l_2$-robustness. In total, 
we improve the previous SOTA for multiple-norm robustness for MNIST from 48.7\% (MSD) to 57.5\% (E-AT fine-tuning of an $l_2$-robust model with 5 epochs) 
which is a significant improvement.
Note that in this case we increase the radii $\epsilon_\infty$ and $\epsilon_1$ to 0.33 and 14 respectively to preserve the $l_2$-robustness: in fact, with such values Eq.~\eqref{eq:bounds_convex} yields $\epsilon_2 \approx 2.16$. However, training from random initialization, in the standard setup, with the larger thresholds leads to worse robustness in the union. We hypothesize that this is due to the increased difficulty of the task to learn: it is known that even single norm adversarial training is problematic when increasing the value of $\epsilon$ \citep{Ding2020MMA}.

\textbf{Experimental details:} For training we use 30 epochs with cyclic learning rate (maximum value 0.05, also used for fine-tuning) and no data augmentation (other settings as for CIFAR-10). As mentioned, we use in adversarial training for multiple norms (SAT, E-AT unif. and E-AT) 50 steps of APGD for $l_1$, and to reduce the training cost we decrease to 5 those for $l_\infty$. Moreover, for training, in $l_1$-APGD we increase the parameter to control the initial sparsity of the updates to 0.1 (default is 0.05). For evaluation, we use the full AutoAttack, since on MNIST FAB \citep{CroHei2019} and Square Attack \citep{ACFH2019square} are at times stronger than PGD-based attacks, as shown in the original papers.

\end{document}